\documentclass[runningheads]{llncs}
\usepackage{graphicx}
\usepackage{comment}
\usepackage{amsmath,amssymb} % define this before the line numbering.
\usepackage{color}
\usepackage{footnote}
\usepackage{multirow}

\usepackage[]{algorithm2e}
\usepackage{subfigure}
\usepackage{cite}
\usepackage{epstopdf}
\usepackage{pifont}
\usepackage[dvipsnames]{xcolor}
\usepackage{caption}
\usepackage{colortbl}
\usepackage{breqn}
\usepackage{booktabs}
\usepackage[export]{adjustbox}

\graphicspath{{figures/}}

\def\figvspace{{}}
\def\tablespace{{}}
\newcommand{\Paragraph}[1]{\noindent \textbf{#1}}

\newcommand{\SubSection}[1]{\subsection{#1}}

\newcommand*{\boxedcolor}{red}
\makeatletter
\renewcommand{\boxed}[1]{\textcolor{\boxedcolor}{%
  \fbox{\normalcolor\m@th$\displaystyle#1$}}}
\makeatother

\makeatletter
\def\hlinew#1{%
  \noalign{\ifnum0=`}\fi\hrule \@height #1 \futurelet
   \reserved@a\@xhline}
\makeatother

\makeatletter
  \newcommand\figcaption{\def\@captype{figure}\caption}
  \newcommand\tabcaption{\def\@captype{table}\caption}
\makeatother

\newenvironment{tight_itemize}{
\begin{itemize}
  \setlength{\topsep}{0pt}
  \setlength{\itemsep}{2pt}
  \setlength{\parskip}{0pt}
  \setlength{\parsep}{0pt}
}{\end{itemize}}

\newcommand\blfootnote[1]{
  \begingroup
  \renewcommand\thefootnote{}\footnote{#1}
  \addtocounter{footnote}{-1}
  \endgroup
}

\definecolor{mygray}{gray}{.7}

\begin{document}
% \linenumbers
\pagestyle{headings}
\mainmatter
\def\ECCVSubNumber{7177}  % Insert your submission number here

\title{Improving Face Recognition from Hard Samples via Distribution Distillation Loss} % Replace with your title

\begin{comment}
\titlerunning{ECCV-20 submission ID \ECCVSubNumber} 
\authorrunning{ECCV-20 submission ID \ECCVSubNumber} 
\author{Anonymous ECCV submission}
\institute{Paper ID \ECCVSubNumber}
\end{comment}
%******************

%\begin{comment}
\titlerunning{Distribution Distillation Loss}

\author{Yuge Huang\inst{1}${}^\ast$ \and Pengcheng Shen\inst{1}${}^\ast$ \and Ying Tai\inst{1}${}^\sharp$ \and Shaoxin Li\inst{1}${}^\sharp$ \and Xiaoming Liu\inst{2} \and Jilin Li\inst{1} \and Feiyue Huang\inst{1} \and Rongrong Ji\inst{3}}
\authorrunning{Y. Huang, P. Shen, and et al.}
\institute{Youtu Lab, Tencent \and
Michigan State University \and
Xiamen University\\
\email{\{yugehuang, quantshen, yingtai, darwinli, jerolinli, garyhuang\}@tencent.com, liuxm@cse.msu.edu, rrji@xmu.seu.cn}}
%\end{comment}
%******************
\maketitle

\begin{abstract}
  Large facial variations are the main challenge in face recognition.
  To this end, previous variation-specific methods make full use of task-related prior to design special network losses, which are typically not general among different tasks and scenarios.
  In contrast, the existing generic methods focus on improving the feature discriminability to minimize the intra-class distance while maximizing the inter-class distance, which perform well on easy samples but fail on hard samples.
  To improve the performance on hard samples, we propose a novel Distribution Distillation Loss to narrow the performance gap between easy and hard samples, which is simple, effective and generic for various types of facial variations.
  Specifically, we first adopt state-of-the-art classifiers such as Arcface to construct two similarity distributions: a teacher distribution from easy samples and a student distribution from hard samples.
  Then, we propose a novel distribution-driven loss to constrain the student distribution to approximate the teacher distribution, which thus leads to smaller overlap between the positive and negative pairs in the student distribution. 
  We have conducted extensive experiments on both generic large-scale face benchmarks and benchmarks with diverse variations on race, resolution and pose. The quantitative results demonstrate the superiority of our method over strong baselines, \textit{e.g.}, Arcface and Cosface. Code will be available at \url{https://github.com/HuangYG123/DDL}.
\end{abstract}

\begin{figure}[t!]
  \centering
  \includegraphics[trim={0 0 0 0mm},clip,width=1\linewidth]{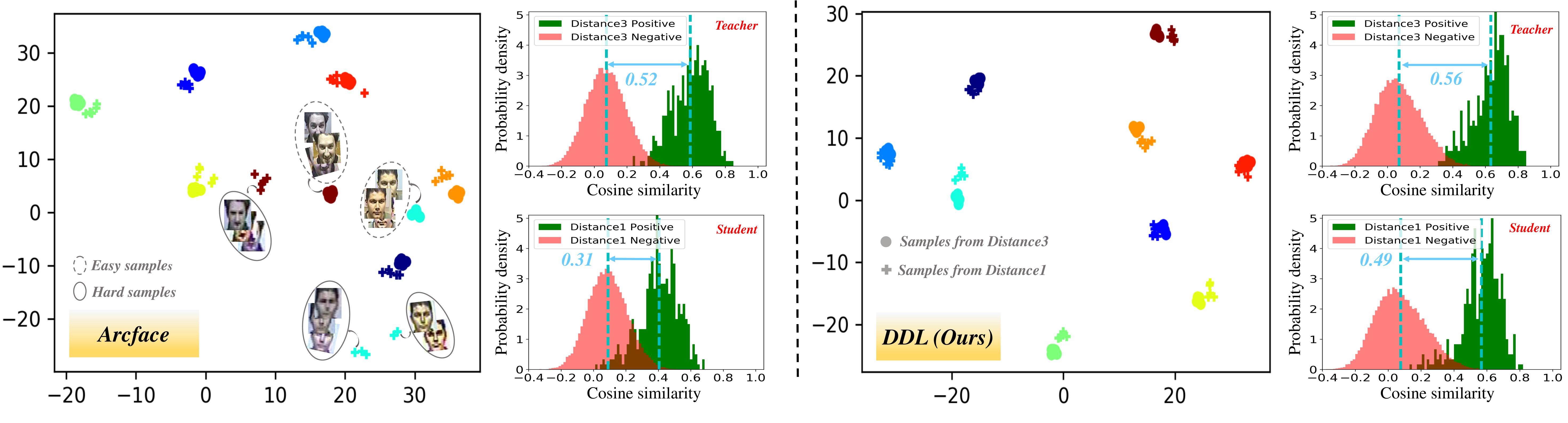}
  \caption{\small \textbf{Comparisons with Arcface~\cite{deng2018Arcface} on SCface~\cite{Grgic2011scface} dataset}.
  T-SNE~\cite{Maaten2008tsne} visualizations on features, where the same color indicates samples of the same subject.
  Distance1 ($d_1$) and Distance3 ($d_3$) indicate low-resolution and high-resolution images, 
  which were captured at distances of $4.2$ and $1.0$m, respectively.
  Each method has two distributions from $d_3$ and $d_1$, where there are also two distributions from the positive and negative pairs with a margin indicating the difference of their expectations. 
 With our distribution distillation loss between the teacher and student distributions, our method effectively narrows the performance gap between the easy and hard samples, decreasing the expectation margin from $\bf{0.21}$ ($0.52$-$0.31$) to $\bf{0.07}$ ($0.56$-$0.49$).
 } 
  \label{fig:comp_f1} \figvspace
\end{figure}

$\blfootnote{$^\ast$ equal contribution.}$
$\blfootnote{$^\sharp$ corresponding author.}$
%%%%%%%%% BODY TEXT
\section{Introduction}
A primary challenge of large-scale face recognition on unconstrained imagery is to handle the diverse variations on pose, resolution, race and illumination, \textit{etc}. 
While some variations are easy to address, many others are relatively difficult.
As in Fig. 1, State-of-the-Art (SotA) facial classifiers like Arcface [6] well address images with small variations with tight groupings in the feature space. We denote these as easy samples.
In contrast, images with large variations are usually far away from the easy ones in the feature space, and are much more difficult to tackle. We denote these as hard samples.
To better recognize these hard samples, there are usually two schemes: {\it variation-specific} and {\it generic} methods.

Variation-specific methods are usually designed for a specific task.
For instance, to achieve pose-invariant face recognition, either handcrafted or learned features are extracted to enhance robustness against pose while remaining discriminative to the identities~\cite{tai2016face}.
Recently, joint face frontalization and disentangled identity preservation are incorporated to facilitate the pose-invariant feature learning~\cite{yin2017towards,tran2017disentangled}.  
To address resolution-invariant face recognition, a unified feature space is learned in~\cite{Sheljar2011fsr,Lei2011local} for mapping Low-Resolution (LR) and High-Resolution (HR) images.
The works~\cite{CT-FSRNet-2018,zhang2018facesr} first apply super-resolution on LR images and then perform recognition on the super-resolved images.
However, the above methods are specifically designed for the respective variations, therefore their ability to generalize from one variation to another is limited.
Yet, it is highly desirable to handle multiple variations in real world recognition systems.

Different from variation-specific methods, generic methods focus on improving the discriminative power of facial features for small intra-class and large inter-class distances. 
Basically, the prior works fall into two categories, \emph{i.e.}, softmax loss-based and triplet loss-based methods.
Softmax loss-based methods regard each identity as a unique class to train the classification networks. 
Since the traditional softmax loss is insufficient to acquire the discriminative features, several variants~\cite{liu2017sphereface,wen2016discriminative,wang2018cosface,deng2018Arcface} are proposed to enhance the discriminability.
In contrast, triplet loss-based methods~\cite{schroff2015facenet,Parkhi2015vggface} directly learn a Euclidean space embedding for each face, where faces from the same person form a separate cluster from faces of other people.
With large-scale training data and well-designed network structures, both types of methods can obtain promising results. 

However, the performance of these methods degrades dramatically on hard samples, such as very large-pose and low-resolution faces.
As illustrated in Fig.~\ref{fig:comp_f1}, 
the features extracted from HR images (\textit{i.e.}, $d_3$) by the strong face classifier of Arcface~\cite{deng2018Arcface} are well separated, but the features extracted from LR images (\textit{i.e.}, $d_1$) cannot be well distinguished. 
From the perspective of the angle distributions of positive and negative pairs, we can easily observe that Arcface exists more confusion regions on LR face images.
It is thereby a natural consequence that such generic methods perform worse on hard samples.

To narrow the performance gap between the easy and hard samples, we propose a novel Distribution Distillation Loss (DDL).
By leveraging the best of both the {\it variation-specific} and {\it generic} methods, our method is generic and can be applied to diverse variations to improve face recognition in hard samples.
Specifically, we first adopt current SotA face classifiers as the baseline (\textit{e.g.}, Arcface) to construct the initial similarity distributions between teacher (\textit{e.g.}, easy samples from $d_3$ in Fig.~\ref{fig:comp_f1}) and student (\textit{e.g.}, hard samples from $d_1$ in Fig.~\ref{fig:comp_f1}) according to the difficulties of samples, respectively. 
Compared to finetuning the baseline models with domain data, 
our method firstly does not require extra data or inference time (\textit{i.e.}, {\it simple});
secondly makes full use of hard sample mining and directly optimizes the similarity distributions to improve the performance on hard samples (\textit{i.e.}, {\it effective});
and finally can be easily applied to address different kinds of large variations in extensive real applications, \textit{e.g.}, women with makeup in fashion stores, surveillance faces in railway stations, and apps looking for missing senior person or children, \textit{etc}. 

To sum up, the contributions of this work are three-fold:
%\vspace{-2mm}
\begin{tight_itemize}
\item Our method narrows the performance gap between easy and hard samples on \textit{diverse} facial variations, which is simple, effective and general.
\item To our best knowledge, it is the first work that adopts similarity distribution distillation loss for face recognition, which provides a new perspective to obtain more discriminative features to better address hard samples.
\item Significant gains compared to the SotA Arcface are reported, $e.g.$, $97.0\%$ over~$92.7\%$ on SCface, $93.4\%$ over~$92.1\%$ on CPLFW, $90.7\%$ over~$89.9\%$ (@FAR=$1e$$-$$4$) on IJB-B and $93.1\%$ over~$92.1\%$ (@FAR=$1e$$-$$4$) on IJB-C.
\end{tight_itemize}
%------------------------------------------------------------------------
\section{Related Work}
%-------------------------------------------------------------------------
%\vspace{-2mm}
\Paragraph{Loss Function in FR.}
Loss function design is pivotal for large-scale face recognition. 
Softmax is commonly used for face recognition~\cite{DeepFace_CVPR14,sun2014deepid,wang2017normface}, which encourages the separability of features but the learned features are not guaranteed to be discriminative.
To address this issue, contrastive~\cite{sun2014deepid2} and triplet~\cite{schroff2015facenet,Parkhi2015vggface} losses are proposed to increase the margin in the Euclidean space.
However, both contrastive and triplet losses occasionally encounter training instability due to the selection of effective training samples. 
As a simple alternative, center loss and its variants~\cite{wen2016discriminative,zhang2017range,deng2017marginal} are proposed to compress the intra-class variance.
More recently, angular margin-based losses~\cite{Wang2018amsoftmax,deng2018Arcface,liu2017sphereface,liu2016large, curricularface-adaptive-curriculum-learning-loss-for-deep-face-recognition} facilitate feature discrimination, and thus lead to larger angular/cosine separability between learned features. 
The above loss functions are designed to apply constraints either between samples, or between sample and center of the corresponding subject. 
In contrast, our proposed loss is {\it distribution} driven.
While being similar to the histogram loss~\cite{Ustinova2016histogram} that constrains the overlap between the distributions of positive and negative pairs across the training set, our loss differs in that we first separate the training set into a teacher distribution (easy samples) and student distribution (hard samples), and then constrain the student distribution to approximate the teacher distribution via our novel loss, which narrows the performance gap between easy and hard samples.

\begin{figure}[t!]
  \centering
  \includegraphics[trim={0 0 0 0mm},clip,width=0.95\linewidth]{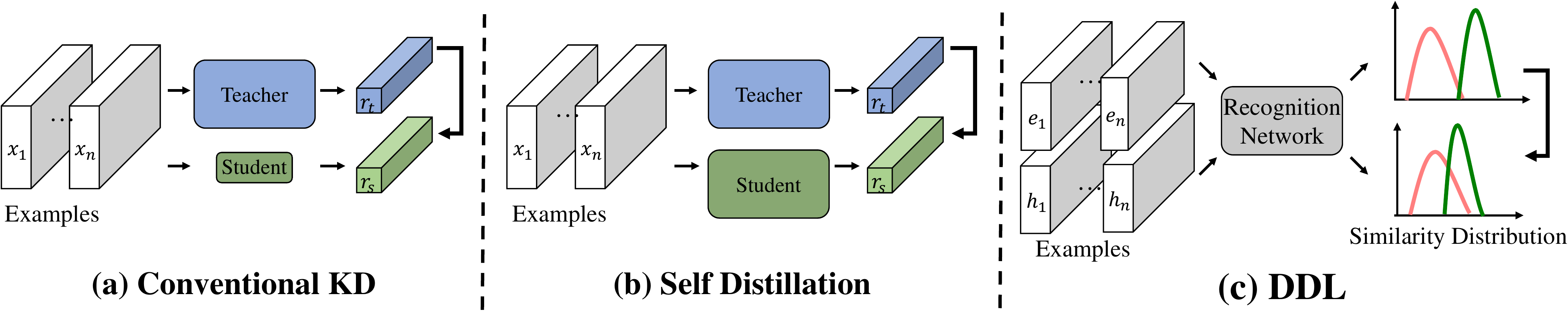}
  %\vspace{-2.5mm}
  \caption{\small \textbf{Comparisons among conventional knowledge distillation, self-distillation and our DDL}. The student in KD is usually smaller than the teacher.
  $\{e\}^n_1$ and $\{h\}^n_1$ indicate the easy and hard samples, respectively.}
  \label{fig:compasison_with_kd} \figvspace
\end{figure}

\Paragraph{Variation-Specific FR.}
Apart from generic solutions~\cite{DeepFace_CVPR14,sun2014deepid} for face recognition, there are also many methods designed to handle specific facial variations, such as resolutions, poses, illuminations, expressions and demographics~\cite{jointly-de-biasing-face-recognition-and-demographic-attribute-estimation}.
For example, 
cross-pose FR~\cite{tran2017disentangled, yin2018multi, zhao20183d,tai2016face} is very challenging, and previous methods mainly focus on either face frontalization or pose invariant representations.
Low resolution FR is also a difficult task, especially in the surveillance scenario.
One common approach is to learn a unified feature space for LR and HR images~\cite{Henning2008facesr,lu2018sr,surveillance}. The other way is to perform super resolution~\cite{DRRN,MemNet_ICCV17,CT-FSRNet-2018} to enhance the facial identity information.
Differing from the above methods that mainly deal with one specific variation, our novel loss is a generic approach to improve FR from hard samples, which is applicable to a wide variety of variations.

\Paragraph{Knowledge Distillation.}
Knowledge Distillation (KD) is an emerging topic. 
Its basic idea is to distill knowledge from a large teacher model into a small one by learning the class distributions provided by the teacher via softened softmax~\cite{Hinton2014kd}.
Typically, Kullback Leibler (KL) divergence~\cite{zhang2018dml,Hinton2014kd} and Maximum Mean Discrepancy (MMD)~\cite{huang2017like} can be adopted to minimize the posterior probabilities between teacher and student models.
More recently, transferring mutual relations of data examples from the teacher to the student is proposed~\cite{park2019rkd, tung2019sp}. In particular, RKD~\cite{park2019rkd} reported that KD can improve the original performance when the student has the same structure as the teacher (\textit{i.e.}, self-distillation). 

Compared to the above distillation methods, our DDL differs in several aspects (see Fig.~\ref{fig:compasison_with_kd}):
% $1$) The inputs of KD/self distillation are randomly selected samples, while inputs of DDL are distinguished into easy and hard samples based on facial variations. 
$1$) KD has at least \textit{two} networks, a teacher and a student, while DDL only learns \textit{one} network. 
Although in KD the student may have the same structure as the teacher (\textit{e.g.}, self-distillation), they have \textit{different parameters} in training. 
$2$) KD uses sample-wise, Euclidean distance-wise or angle-wise constraints, while DDL proposes a novel cosine similarity distribution-wise constraint which is specifically designed for face recognition. 
$3$) To our best knowledge, currently no KD methods outperform SotA face classifiers on face benchmarks, while DDL consistently outperforms the SotA Arcface classifier.

\begin{figure}[t!]
  \centering
  \includegraphics[trim={0 0 0 0mm},clip,width=0.92\linewidth]{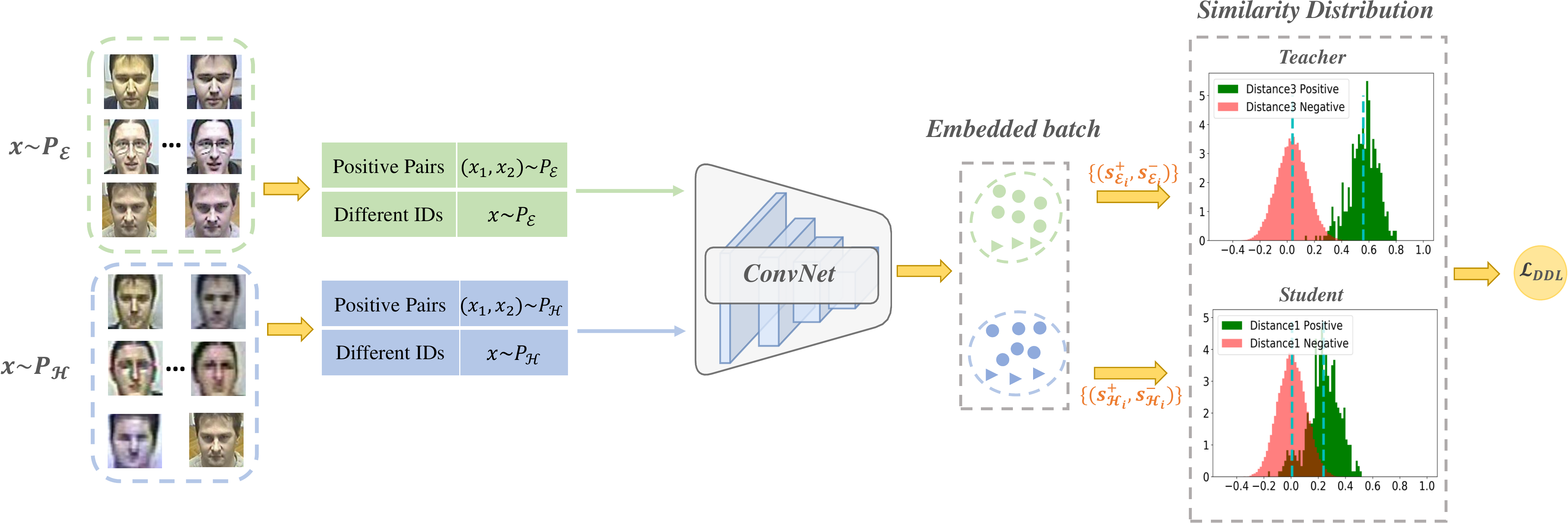}
  %\vspace{-2mm}
  \caption{\small \textbf{Illustration of our DDL}.
  We sample $b$ positive pairs (\textit{i.e.}, $2b$ samples) and $b$ samples with different identities, for both the teacher $P_{\mathcal{E}}$ and student $P_{\mathcal{H}}$ distributions, to form one mini-batch (\textit{i.e.}, $6b$ in total).
  $\{(s^+_{\mathcal{E}_i},s^-_{\mathcal{E}_i}) |i=1,...,b\}$ indicates 
  we construct $b$ positive and negative pairs from $P_{\mathcal{E}}$ via Eqs.~\ref{eq:s+} and~\ref{eq:s-} respectively to estimate the teacher distribution.
  $\{(s^+_{\mathcal{H}_i},s^-_{\mathcal{H}_i}) | i=1,...,b\}$ also indicates we construct $b$ positive and negative pairs from $P_{\mathcal{H}}$ via Eqs.~\ref{eq:s+} and~\ref{eq:s-} respectively to estimate the student distribution.
  } 
  \label{fig:framework} \figvspace
\end{figure}

%------------------------------------------------------------------------
\section{The Proposed Method}
Fig.~\ref{fig:framework} illustrates the framework of our DDL.
We separate the training set into two parts, \emph{i.e.}, $\mathcal{E}$ for easy samples and $\mathcal{H}$ for hard samples to form the teacher and student distributions, respectively.
In general, for each mini-batch during training, we sample from both parts.
To ensure a good teacher distribution, we use the SotA FR model~\cite{deng2018Arcface} as our initialization.
The extracted features are used to construct the positive and negative pairs (Sec.~\ref{sec:3.1}), which are further utilized to estimate the similarity distributions (Sec.~\ref{sec:3.2}).
Finally, based on the similarity distributions, the proposed DDL is utilized to train the classifier (Sec.~\ref{sec:3.3}).  
\SubSection{Sampling Strategy from $P_{\mathcal{E}}$ and $P_{\mathcal{H}}$}
\label{sec:3.1}
First, we introduce the details on how we construct the positive and negative pairs in one mini-batch during training.
Given two types of input data from both $P_{\mathcal{E}}$ and $P_{\mathcal{H}}$,
each mini-batch consists of four parts, two kinds of positive pairs (\textit{i.e.}, $(x_{1}, x_{2})$ $\sim$ $P_{\mathcal{E}}$ and $(x_{1}, x_{2})$ $\sim$ $P_{\mathcal{H}}$), and two kinds of samples with different identities (\textit{i.e.}, $x$ $\sim$ $P_{\mathcal{E}}$ and $x$ $\sim$ $P_{\mathcal{H}}$).
To be specific, we on one hand construct $b$ positive pairs (\textit{i.e.}, $2b$ samples), and on the other hand $b$ samples with different identities from $P_{\mathcal{E}}$ and $P_{\mathcal{H}}$.
As the result, there are $6b=(2b+b)*2$ samples in each mini-batch (see Fig.~\ref{fig:framework} for more details).

\Paragraph{Positive Pairs.}
The positive pairs are constructed offline in advance, and each pair consist of two samples with the same identity. 
As shown in Fig.~\ref{fig:framework}, samples of each positive pair are arranged in order. 
After embedding data into a high-dimensional feature space by a deep network $\mathcal{F}$, the similarity of a positive pair $s^{+}$ can be obtained as follows:
\begin{equation}
\label{eq:s+}
\small
    s^{+}_{i} = <\mathcal{F}(x_{pos_{i1}}), \mathcal{F}(x_{pos_{i2}})>, i=1,...,b
\end{equation}
where $x_{pos_{i1}}$, $x_{pos_{i2}}$ are the samples of one positive pair. 
Note that positive pairs with similarity less than $0$ are usually outliers, which are deleted as a practical setting since our main goal is not to specifically handle noise.

\Paragraph{Negative Pairs.}
Different from the positive pairs, we construct negative pairs online from the samples with different identities via hard negative mining, which selects negative pairs with the largest similarities. 
To be specific, the similarity of a negative pair $s^{-}$ is defined as:
\begin{equation}
\label{eq:s-}
\small
    s^{-}_{i} = \max_j \Big(\{s^{-}_{ij}=<\mathcal{F}\big(x_{neg_i}\big), \mathcal{F}(x_{neg_j})>| j=1,...,b\}
    \Big), 
\end{equation}
where $x_{neg_i}$, $x_{neg_j}$ are from different subjects.
Once the similarities of positive and negative pairs are constructed, the corresponding distributions can be estimated, which is described in the next subsection.

\SubSection{Similarity Distribution Estimation}
\label{sec:3.2}
The process of similarity distribution estimation is similar to~\cite{Ustinova2016histogram}, which is performed in a simple and piece-wise differentiable manner using $1$D histograms with soft assignment. % operations.
Specifically, two samples $x_{i}$, $x_{j}$ from the same person form a positive pair, and the corresponding label is denoted as $m_{ij}=+1$. 
In contrast, two samples from different persons form a negative pair, and the label is denoted as $m_{ij} = -1$.
Then, we obtain two sample sets $\mathcal{S}^{+} = \{s^{+} = \langle \mathcal{F}(x_{i}), \mathcal{F}(x_{j}) \rangle | m_{ij} = + 1\}$ and $\mathcal{S}^{-} = \{s^{-} = \langle \mathcal{F}(x_{i}), \mathcal{F}(x_{j}) \rangle | m_{ij} = - 1\}$ corresponding to the similarities of positive and negative pairs, respectively.

Let $p^+$ and $p^-$ denote the two probability distributions of $\mathcal{S}^{+}$ and  $\mathcal{S}^{-}$, respectively.
As in cosine distance-based methods~\cite{deng2018Arcface}, the similarity of each pair is bounded to $[-1,1]$, which is demonstrated to simplify the task~\cite{Ustinova2016histogram}.  
Motivated by the histogram loss, we estimate this type of one-dimensional distribution by fitting simple histograms with uniformly spaced bins.
We adopt $R$-dimensional histograms $H^+$and $H^-$, with the nodes $t_{1}=-1$, $t_{2}$, $\cdots$, $t_{R}=1$ uniformly filling $[-1,1]$ with the step $\triangle = \frac{2}{R-1}$.  
Then, we estimate the value $h_{r}^{+}$ of the histogram $H^{+}$ at each bin as: 
\begin{equation}\small
    h_{r}^{+}=\frac{1}{|S^+|}\sum_{(i,j):m_{ij=+1}}\delta_{i,j,r},
\end{equation}
where $(i, j)$ spans all the positive pairs. 
Different from \cite{Ustinova2016histogram}, the weights $\delta_{i,j,r}$ are chosen by an exponential function as:
\begin{equation}\small
    \delta_{i,j,r} = exp(-\gamma{(s_{ij} - t_{r})^{2}}),
\end{equation}
where $\gamma$ denotes the spread parameter of Gaussian kernel function, and $t_{r}$ denotes the ${r}$th node of histograms. We adopt the Gaussian kernel function because it is the most commonly used kernel function for density estimation and robust to the small sample size.
The estimation of $H^{-}$ proceeds analogously. 

\begin{figure}[t!]
  \centering
  \includegraphics[trim={0 5 0 0mm},clip,width=0.85\textwidth]{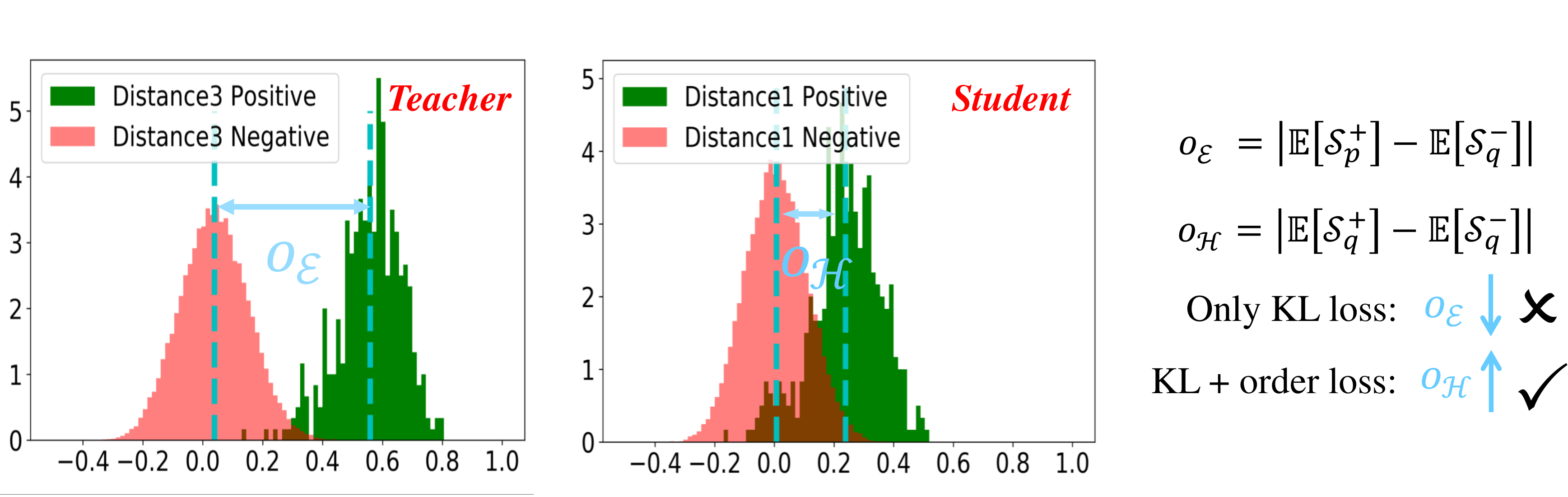}
  %\vspace{-4mm}
  \caption{\small \textbf{Illustration of the effects of our order loss}.
  Similarity distributions are constructed by Arcface~\cite{deng2018Arcface} on SCface, in which we have $2$ kinds of order distances formed from both of the teacher and student distributions according to Eq.~\ref{eq:order_loss}.
  } 
  \label{fig:order} \figvspace
\end{figure}

\SubSection{Distribution Distillation Loss} 
\label{sec:3.3}
We make use of SotA face recognition engines like~\cite{deng2018Arcface}, to obtain the similarity distributions from two kinds of samples: easy and hard samples. Here, easy samples indicate that the FR engine performs well, in which the similarity distributions of positive and negative pairs are clearly separated (see the teacher distribution in Fig.~\ref{fig:order}), while hard samples indicate that the FR engine performs poorly, in which the similarity distributions may be highly overlapped (see the student distribution in Fig.~\ref{fig:order}).

\Paragraph{KL Divergence Loss.}
To narrow the performance gap between the easy and hard samples, we constrain the similarity distribution of hard samples (\textit{i.e.}, student distribution) to approximate the similarity distribution of easy samples (\textit{i.e.}, teacher distribution).
The teacher distribution consists of two similarity distributions of both positive and negative pairs, denoted as $P^{+}$ and $P^{-}$, respectively. 
Similarly, the student distribution also consists of two similarity distributions, denoted as $Q^{+}$ and $Q^{-}$. 
Motivated by the previous KD methods~\cite{Hinton2014kd,zhang2018dml}, we adopt the KL divergence to constrain the similarity between the student and teacher distributions, which is defined as follows:
\begin{equation}
\small
\begin{aligned}
    &\mathcal{L}_{KL} = \lambda_{1}{\mathbb{D}_{KL}}(P^+||Q^+) + \lambda_{2}{\mathbb{D}_{KL}}(P^-||Q^-) \\
    &=\underbrace{\lambda_{1}\sum_s P^+(s)\log\frac{P^+(s)}{Q^+(s)}}_{KL\,loss\,on\,pos.\,pairs} + \underbrace{\lambda_{2}\sum_s P^-(s)\log\frac{P^-(s)}{Q^-(s)}}_{KL\,loss\,on\,neg.\,pairs},
\end{aligned}
\end{equation}
where $\lambda_{1}$, $\lambda_2$ are the weight parameters.

\Paragraph{Order Loss.}
However, only using KL loss does not guarantee good performance. In fact, the teacher distribution may choose to approach the student distribution and leads to more confusion regions between the distributions of positive and negative pairs, which is the opposite of our objective (see Fig.~\ref{fig:order}). 
To address this problem, we design a simple yet effective term named \textit{order loss}, which minimizes the distances between the expectations of similarity distributions from the negative and positive pairs to control the overlap. 
Our order loss can be formulated as follows:
\begin{equation}\small
\label{eq:order_loss}
    \mathcal{L}_{order} = -\lambda_{3}\sum_{(i,j)\in{(p,q)}}( \mathop{\mathbb{E}}[{\mathcal{S}_{i}^{+}}] - \mathop{\mathbb{E}}[{\mathcal{S}_{j}^{-}}]),
\end{equation}
where $\mathcal{S}_{p}^{+}$ and $S_{p}^{-}$ denote the similarities of positive and negative pairs of the teacher distribution; $\mathcal{S}_{q}^{+}$ and $S_{q}^{-}$ denote the similarities of positive and negative pairs of the student distribution; and $\lambda_{3}$ is the weight parameter.

In summary, the entire formulation of our distribution distillation loss is: $\mathcal{L}_{DDL} = \mathcal{L}_{KL} + \mathcal{L}_{order}$.
DDL can be easily extended to multiple student distributions varied from one specific variation as follows:
\begin{equation}
\label{eq:multi_ddl}
\small
  %\resizebox{0.9\hsize}{!} {$
    \mathcal{L}_{DDL} =
    \sum\limits_{i=1}^{K}\mathop{\mathbb{D}_{KL}}(P||Q_i)
    - \lambda_{3}\sum_{i, j\in (p, q_{1}...q_{K})}( \mathop{\mathbb{E}}[{\mathcal{S}_{i}^{+}}] - \mathop{\mathbb{E}}[{\mathcal{S}_{j}^{-}}]),
  %}
\end{equation}
where $K$ is the number of student distributions.
Further, to maintain the performance on easy samples, we incorporate the loss function of Arcface~\cite{deng2018Arcface}, and thus the final loss is: 
\begin{equation}\small
    \mathcal{L}(\Theta) = \mathcal{L}_{DDL} + \mathcal{L}_{Arcface},
\end{equation}
where $\Theta$ denotes the parameter set.
Note that $\mathcal{L}_{Arcface}$ can be easily replaced by any kind of popular losses in FR.

\SubSection{Generalization on Various Variations} \label{sec:3.4}
Next, we discuss the generalization of DDL on various variations, which defines our \textit{application scenarios} and how we select easy/hard samples.
Basically, we can distinguish the easy and hard samples according to whether the image contains large facial variations that may hinder the identity information, \textit{e.g.}, low-resolution and large pose variation.  

\begin{figure}[t!]
  \begin{minipage}[b]{0.56\linewidth}
     \centering
     \includegraphics[trim={0 0 0 0mm},clip,width=0.98\linewidth,left]{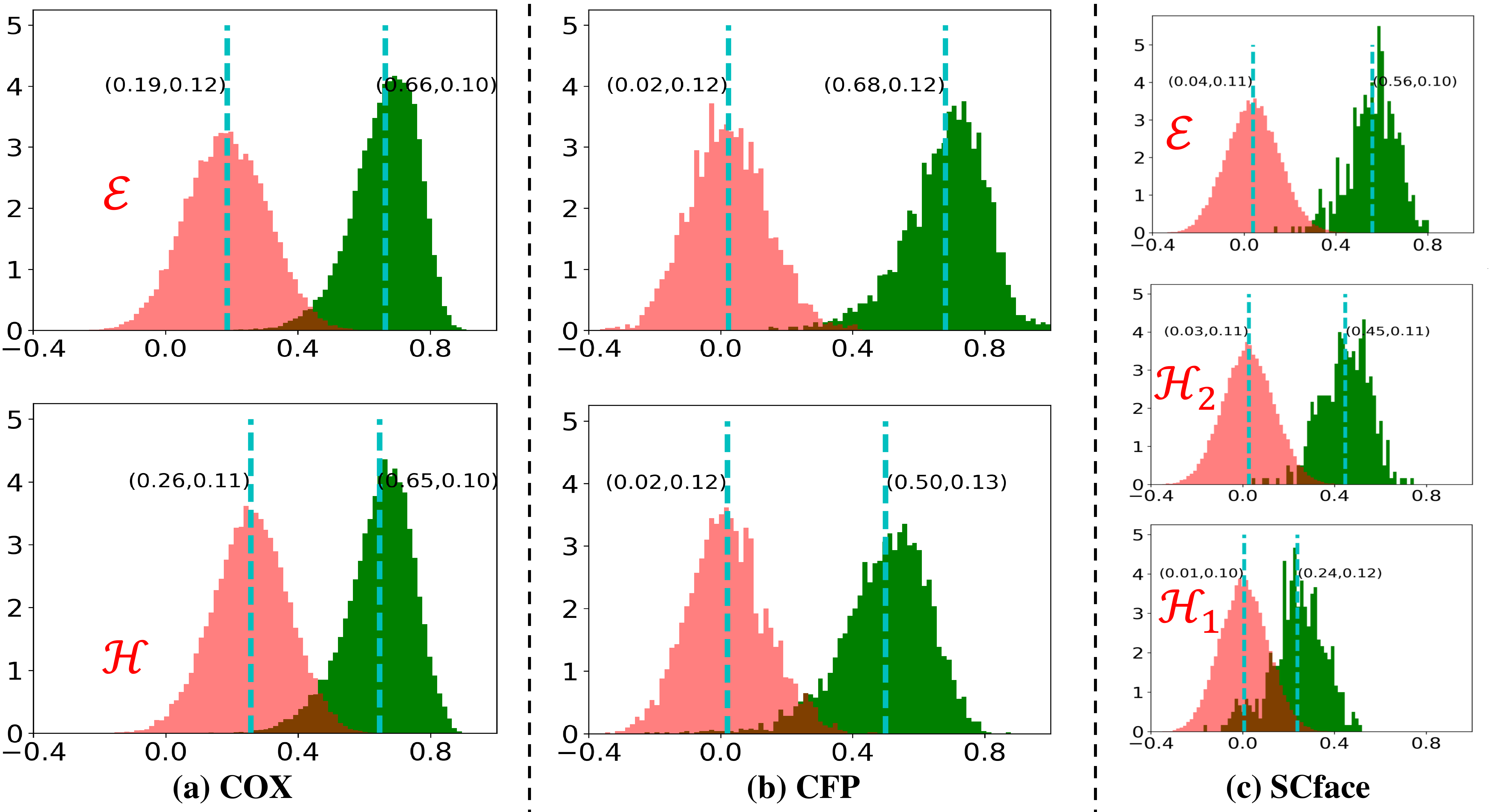}
      %\vspace{-2mm}
      \caption{\footnotesize \textbf{Similarity distribution differences} between easy and hard samples on various variations, including race on COX, pose on CFP, and resolution on SCface respectively. ($\cdot$,$\cdot$) indicates the mean and standard deviation.}
      \label{fig:scenarios}
  \end{minipage}
  %\hspace{1.5mm}
  \begin{minipage}[b]{0.415\linewidth}
     \centering
     %\vspace{-4.5mm}
     \includegraphics[trim={0 0 0 0mm},clip,width=0.98\linewidth,right]{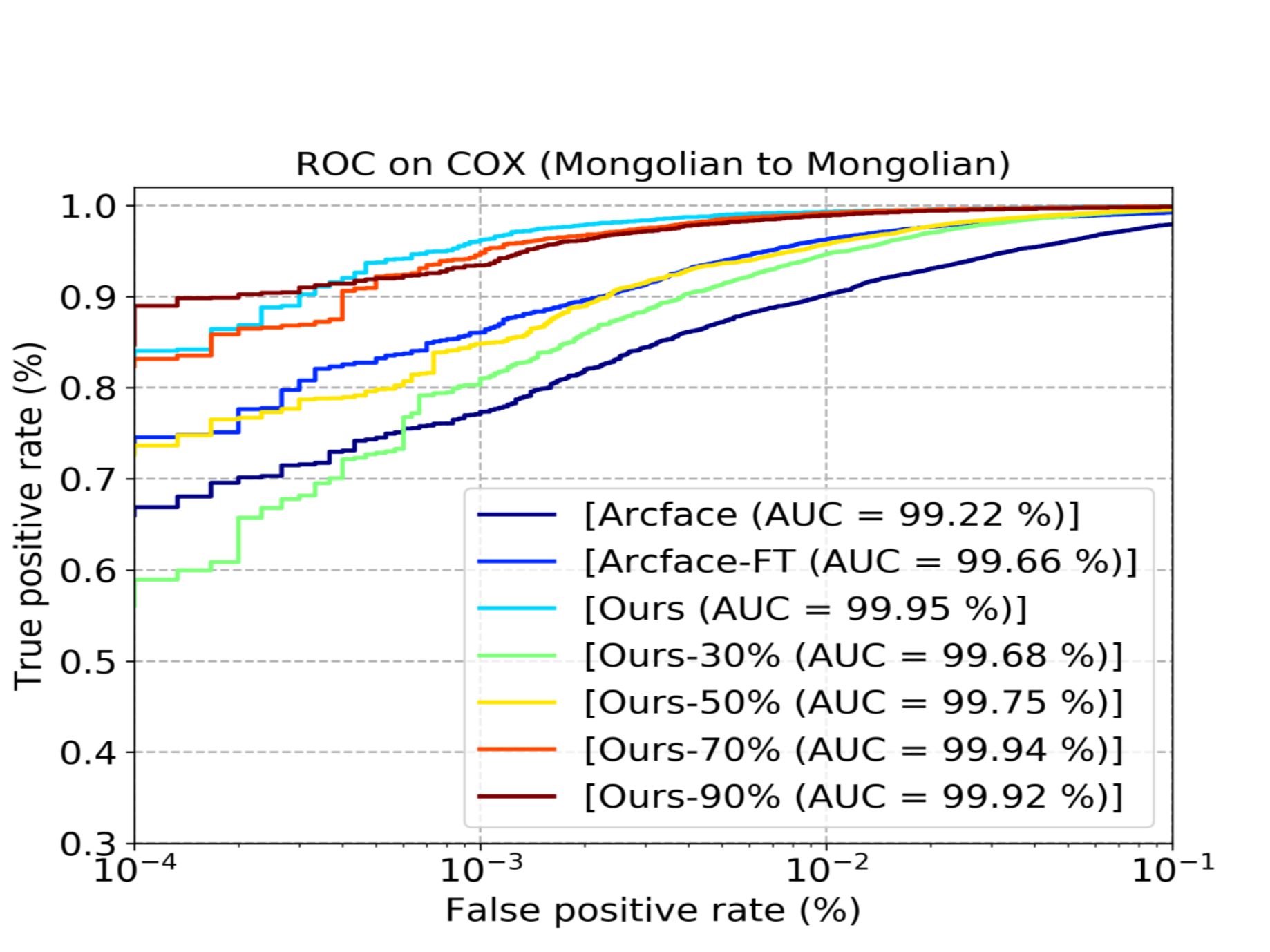}
     %\vspace{-1mm}
     \caption{\footnotesize \textbf{Effects of number of training subjects} on COX. Compared to Arcface-FT, DDL achieves comparable results with only \textit{half} the number of training subjects.} 
     \label{fig:ablation_subjects}
  \end{minipage}
  \figvspace
\end{figure}

\Paragraph{Observation from Different Variations.}
Our method assumes that two or more distributions, each computed from a subset of training data, have differences among themselves, which is a popular phenomenon in face recognition and is demonstrated in Fig.~\ref{fig:scenarios}.
It shows similarity distributions of normal and challenging samples based on Arcface~\cite{deng2018Arcface} trained on CASIA except CFP, which is trained on VGGFace$2$.
As we can see,
$1$) since CASIA is biased to Caucasian, Mongolian samples in COX are more difficult and thus relatively regarded as the hard samples,
$2$) different variations share a common observation that the similarity distributions of challenging samples are usually different from those of easy samples,
$3$) variations with different extents may have different similarity distributions ($e.g.$, $\mathcal{H}$$1$ and $\mathcal{H}$$2$ in Fig.~\ref{fig:scenarios}(c)). 
In summary, when a task satisfies that \textit{the similarity distributions differ between easy and hard samples}, our method is a good solution and one can enjoy the performance improvement by properly constructing the positive and negative pairs, as validated in Sec.~\ref{compare_SotA}.

\Paragraph{Performance Balance Between Easy and Hard Samples.}
Improving the performance on hard samples while maintaining the performance on easy samples is a trade-off. 
Two factors in our method help maintain performance on easy samples. 
First, we incorporate the SotA Arcface loss~\cite{deng2018Arcface} to maintain feature discriminability on easy samples. 
Second, our order loss minimizes the distance between the expectations of similarity distributions from the negative and positive pairs, which helps control the overlap between positive and negative pairs. 

\Paragraph{Discussions on Mixture Variations.}
As shown in Eq.~\ref{eq:multi_ddl}, our method can be easily extended to multiple variations for one task (\textit{e.g.}, low resolution, large pose, \textit{etc}).
An alternative is to mix the variations with different extents from one task into $one$ student distribution, which, as shown in Sec.~\ref{sec:abl}, is not good enough to specifically model the different extents and tends to lead to lower performance.
As for different variations from different tasks, one may also construct multiple teacher-student distribution pairs to address the corresponding task respectively, which can be a good future direction.

\section{Experiments}
\subsection{Implementation Details} 
\Paragraph{Datasets.}
We separately employ SCface~\cite{Grgic2011scface}, COX~\cite{Huang2015cox}, CASIA-WebFace~\cite{Yi2014learning}, VGGFace$2$~\cite{cao2018vggface2} and the refined MS$1$M~\cite{deng2018Arcface} as our training data to conduct fair comparisons with other methods.
We extensively test our method on benchmarks with diverse variations, \textit{i.e.}, COX on race, SCface on resolution, CFP and CPLFW on Pose, as well as generic large-scale benchmarks IJB-B and IJB-C.
For COX, the data are collected from two races: Caucasian and Mongolian. Since no race label is given, 
we manually label $428$ Mongolians and $572$ Caucasians to conduct experiments, in which 
half of both races are used for finetuning and the others for testing.
For SCface, following~\cite{lu2018sr}, $50$ subjects are used for finetuning and $80$ subjects are for testing. 
In the testing stage, we conduct face identification, where HR image is used as the gallery and LR images with three different resolutions form the probe.
Specifically, the LR images are captured at three distances: $4.2$m for $d_1$, $2.6$m for $d_2$ and $1.0$m for $d_3$. 
We split easy and hard samples according to the main variation in each dataset. For race, since the dataset on which the model is pre-trained is biased to Caucasian, Mongolian samples on COX  are  more  difficult  and  thus  relatively regarded as the hard samples. For pose, we estimate the pose of each image \cite{Ruiz_2018_CVPR_Workshops} on VGGFace$2$, and images with yaw $<10^{\circ}$ and yaw $>45^{\circ}$ are used as easy and hard samples respectively. For resolution, images captured under $d_3$ and $d_1$/$d_2$ are used as easy and hard samples respectively on SCface.

\Paragraph{Training Setting.}
We follow~\cite{deng2018Arcface,wang2018cosface} to generate the normalized faces ($112\times 112$) with five landmarks~\cite{zhang2016mtcnn}.
For the embedding network, we adopt ResNet$50$ and ResNet$100$ as in~\cite{deng2018Arcface}.
Our work is implemented in Tensorflow~\cite{tensorflow2015}.
We train models on $8$ NVIDIA Tesla P$40$ GPUs.
On SCface, we set the number of positive/negative pairs as $b=16$, thus the batch size on one GPU is $3b\times3=144$, including one teacher distribution and two student distributions (see Fig.~\ref{fig:scenarios}(c)). 
On other datasets, we set $b$ to $32$, thus the batch size per GPU is $3b\times2=192$. 
The numbers of iterations are $1$K, $2$K and $20$K on SCface, COX and VGGFace2, respectively. 
The models are trained with SGD, with momentum $0.9$ and weight decay $5e^{-4}$. 
The learning rate is $1e^{-3}$, and is divided by $10$ at half of iterations. All of the weight parameters are consistent across all the experiments. $\lambda_1$, $\lambda_2$ and $\lambda_3$ are set to $1e^{-1}$, $2e^{-2}$ and $5e^{-1}$, respectively. 
% $\lambda_1$/$\lambda_2$ depend on the positive/negative distribution gap between easy and hard samples respectively, while $\lambda_3$ depends on the gap between positive and negative distributions. 
% All weights are set based on the scales of loss terms, and we set a larger $\lambda_1$ to focus on minimizing the gap of positive distributions.

\subsection{Ablation Study} \label{sec:abl}
\Paragraph{Effects of Distance Metric on Distributions.}
We investigate the effects of several commonly used distribution metrics to constrain the teacher and student distributions in our DDL, including KL divergence, Jensen-Shannon (JS) divergence, and Earth Mover Distance (EMD).
Although KL divergence does not qualify as a statistical metric, it is widely used as a measure of how one distribution is different from another. 
JS divergence is a symmetric version of KL divergence. 
EMD is another distance function between distributions on a given metric space and has seen success in image synthesis~\cite{GAN_NIPS14}. 
We incorporate our order loss with the above distance metrics,
and report the results in Tab.~\ref{tab:SCface_distance}.
We choose KL divergence in our DDL since it achieves the best performance, which shares similar conclusion with~\cite{zhang2018dml}.
To further investigate the effectiveness of each component in our loss, we train the network with each component separately. 
As shown in Tab.~\ref{tab:SCface_distance}, only KL or only Order does not guarantee satisfying performance, while using both components leads to better results. 

\Paragraph{Effects of Random vs.~Hard Mining.}
To investigate the effect of hard sample mining in our method, we train models on SCface with the corresponding strategy (\textit{i.e.}, negative pairs with the largest similarity are selected), and without the strategy by randomly selecting the negative pairs, respectively. 
The comparative results are reported in Tab.~\ref{tab:SCface_distance}. 
Comparing with the results of ``Random" selecting, it is clear that our hard mining version outperforms the one without.

\Paragraph{Effects of Mixture vs.~Specific training.}
As mentioned in Sec.~\ref{sec:3.4}, we basically construct different student distributions for samples with different extents of variations on SCface.
Here, we mix two variations from $d_1$ and $d_2$ into \textit{one} student distribution.
The comparison between our specific and mixture training is also shown in Tab.~\ref{tab:SCface_distance}.
As we expected, the mixture version is worse than the specific version, but is still better than the conventional finetuning ($i.e.$, Avg. being $86.3$), which indicates that properly constructing different hard samples for the target tasks may maximize the advantages of our method.

\Paragraph{Effects of Number of Training Subjects.}
Here, we conduct tests on COX dataset to show the effects of using different numbers of training subjects.
Specifically, we adopt $10\%$, $30\%$, $50\%$, $70\%$, $90\%$ and $100\%$ of training subjects, respectively.
A pre-trained Arcface on CASIA is used as the baseline.
For fair comparison, we also compare our method against Arcface with conventional finetuning (\textit{i.e.}, Arcface-FT).
From Fig.~\ref{fig:ablation_subjects} we see that:
$1$) Compared to Arcface-FT, our method clearly boosts the performance on Mongolian-Mongolian verification tests with comparable training data. 
$2$) Our method can have comparable performance with the only \textit{half} of the entire training subjects, which demonstrate the superiority of utilizing the global similarity distributions.
\begin{table*}[t!]
\begin{center}
\small
\caption{\small \textbf{Extensive ablation studies on SCface dataset. 
All methods are trained on \textbf{CASIA} with a \textbf{ResNet$50$} backbone.}
% $d_1$, $d_2$ and $d_3$ indicate images in SCface~\cite{Grgic2011scface} captured at distances of $4.2$, $2.6$ and $1.0$m, respectively. 
Each color corresponds to a type of ablation study experimental setting.
}
\label{tab:SCface_distance}
\resizebox{0.98\textwidth}{!}{
\begin{tabular}{c|cccc|cc|cc|ccc|c}
\hlinew{1.2pt}
%\hline
& EMD & JS & KL & Order & Random & Hard Mining & Mixture & Specific &  $d_1$ & $d_2$ & $d_3$ & Avg. \\ \hline
Arcface-FT
& &  &  &  &  &  &  &  &  $67.3$ & $93.5$ & $98.0$ & $86.3$ \\ \hline
\multirow{4}{*}{Distance metric} 
& \cellcolor{mygray}\checkmark & \cellcolor{mygray} & \cellcolor{mygray} & \cellcolor{mygray} \checkmark  &  & \checkmark  & & \checkmark & $78.0$ & $97.8$ & $96.8$ & $90.5$ \\
& \cellcolor{mygray}& \cellcolor{mygray}\checkmark &  \cellcolor{mygray}& \cellcolor{mygray}\checkmark &  & \checkmark  & & \checkmark & $83.0$ & $\bf 98.3$ & $\bf 99.0$ & $93.4$ \\
& \cellcolor{mygray}& \cellcolor{mygray} &  \cellcolor{mygray}\checkmark & \cellcolor{mygray} &  & \checkmark & & \checkmark    &  $76.0$ & $94.3$ & $98.5$ & $89.6$ \\
& \cellcolor{mygray}& \cellcolor{mygray} & \cellcolor{mygray} & \cellcolor{mygray}\checkmark &  & \checkmark  & & \checkmark &  $80.8$ & $97.5$ & $\bf 99.0$ & $92.4$ \\ \hline
Mining strategy & 
&  & \checkmark & \checkmark & \cellcolor{green}\checkmark &\cellcolor{green}  & & \checkmark & $80.3$ & $96.3$ & $95.3$ & $90.6$ \\ \hline\
Mixture training &
&  & \checkmark & \checkmark &  & \checkmark & \cellcolor{blue}\checkmark & \cellcolor{blue} & $81.5$ & $97.0$ & $97.8$ & $92.1$ \\ \hline \hline
DDL (ours)
& &  & \checkmark & \checkmark &  & \checkmark &  & \checkmark & $\bf 86.8$ & $\bf 98.3$ & $98.3$ & $\bf 94.4$ \\ 
\hline
\end{tabular}
}
\tablespace
\end{center}
\end{table*}

\subsection{Comparisons with SotA Methods} \label{compare_SotA}
\Paragraph{Resolution on SCface.}
% SCface defines face identification with HR and LR faces. 
SCface mimics the real-world surveillance watch-list problem, where the gallery contains HR faces and the probe consists of LR faces captured from surveillance cameras.
We compare our method with SotA low-resolution face recognition methods in Tab.~\ref{tab:SCface}.
Most results are directly cited from~\cite{lu2018sr}, while the results of Arcface come from  our re-implementation.
From Tab.~\ref{tab:SCface}, we have some observations:
$1$) The baseline Arcface achieves much better results than the other methods without finetuning, especially on the relatively high-resolution images from $d_3$.
$2$) Our (CASIA+ResNet$50$)-FT version already outperforms all of the other methods, including Arcface (MS$1$M+ResNet$100$)-FT, which uses a larger model that is trained by a much larger dataset.
$3$) We achieve significant improvement on $d_1$ setting, which is the hardest. 
This demonstrates the effectiveness of our novel loss.
$4$) Histogram loss performs poorly, which demonstrates the effects of our constraint between teacher and student distributions.

Moreover, different to the prior hard mining methods~\cite{schroff2015facenet,lin2017focal,shrivastava2016training} where the hard samples are mined based on the loss values during the training process, we \textit{pre-define} hard samples according to human prior.
Penalizing \textit{individual} samples or triplets as in previous hard mining methods does not leverage sufficient contextual insight of the overall distribution.
DDL minimizes the difference of \textit{global} similarity distributions between the easy and hard samples, which is more robust for tackling hard samples and against the noisy samples.
The word "global" means our method leverages sufficient contextual insight of the overall distribution in a mini-batch, rather than focusing on a sample.

\setcounter{footnote}{0}
  \begin{savenotes}
  \begin{table}[t!]
  \begin{minipage}[ht]{0.475\linewidth}
  \small
  \caption{\footnotesize \textbf{Rank-$1$ performance (\%) of face identification on SCface testing set}. '-FT' represents finetuning with training set from SCface. }
  \label{tab:SCface}
  \resizebox{0.965\columnwidth}{!}{
  \begin{tabular}{l|ccc|c}
  \hlinew{1.2pt}
  Distance $\to$ & $d_1$ & $d_2$ & $d_3$ & avg. \\ \hline
  LDMDS~\cite{yang2018discriminative} & $62.7$ & $70.7$ & $65.5$ & $66.3$ \\
  Center Loss~\cite{wen2016discriminative} & $36.3$ & $81.8$ & $94.3$ & $70.8$ \\
  Arcface (CASIA+R$50$) & $48.0$ & $92.0$ & ${99.3}$ & $79.8$ \\
  Arcface (MS$1$M+R$100$) & $58.9$ & $98.3$ & $\bf{99.5}$ & $85.5$ \\ \hline
  Center Loss-FT & $54.8$ & $86.3$ & $95.8$ & $79.0$ \\
  DCR-FT~\cite{lu2018sr} & $73.3$ & $93.5$ & $98.0$ & $88.3$ \\
  Histogram (CASIA+R$50$)-FT~\cite{Ustinova2016histogram} & $74.3$ & $95.0$ & $97.3$ & $88.8$ \\
  OHEM (CASIA+R$50$)-FT~\cite{shrivastava2016training} &  $82.5$ & $97.3$ & $97.5$ & $92.7$\\
  Focal (CASIA+R$50$)-FT~\cite{lin2017focal} & $76.8$ & $95.5$ & $96.8$ & $89.7$ \\
  Triplet (CASIA+R$50$)-FT~\cite{deng2018Arcface} & $84.2$ & $97.2$ & $99.2$ & $93.5$\\
  Arcface (CASIA+R$50$)-FT & $67.3$ & $93.5$ & $98.0$ & $86.3$ \\
  Arcface (MS$1$M+R$100$)-FT & $80.5$ & $98.0$ & ${99.5}$ & $92.7$ \\ \hline\hline
  \bf Ours (CASIA+R$50$) & ${86.8}$ & ${98.3}$ & ${98.3}$ & ${94.4}$ \\ 
  \bf Ours (MS$1$M+R$100$) & $\bf{93.2}$ & $\bf{99.2}$ & ${98.5}$ & $\bf{97.0}$  \\
  \hline
  \end{tabular}
  }
  \end{minipage}
  %\hspace{2mm}
  \begin{minipage}[ht]{0.475\linewidth}
  \small
  \caption{\footnotesize \textbf{Verification comparisons with SotA methods} on LFW and two popular pose benchmarks, including CFP-FP and CPLFW.}
  \label{tab:comp_pose}
  \resizebox{1\columnwidth}{!}{
  \begin{tabular}{l|ccc}
  \hlinew{1.2pt}
  Methods (\%) & LFW & CFP-FP & CPLFW \\ \hline
  Triplet Loss (CVPR'$15$) & $98.98$ & $91.90$ & $-$ \\
  Center Loss (ECCV'$16$)~\cite{wen2016discriminative} & $98.75$ & $-$ & $77.48$ \\
  SphereFace (CVPR'$17$)~\cite{liu2017sphereface} & $99.27$ & $-$ & $81.40$\\
  DRGAN (CVPR'$17$)~\cite{tran2017disentangled} & $-$ & $93.41$ & $-$ \\
  Peng~\textit{et al.} (ICCV'$17$)~\cite{peng2017rec} & $-$ & $93.76$ & $-$ \\
  Yin~\textit{et al.}  (TIP'$17$)~\cite{yin2018multi} & $98.27$ & $94.39$ & $-$ \\
  VGGFace2 (FG'$18$)~\cite{cao2018vggface2} & $99.43$ & $-$ & $84.00$\\
  Dream (CVPR'$18$)~\cite{cao2018pose} & $-$ & $93.98$ & $-$\\
  Deng~\textit{et al.} (CVPR'$18$)~\cite{deng2018uvgan} & $99.60$ & $94.05$ & $-$\\ 
  SV-Arc-Softmax (arXiv'$19$)~\cite{wang2018support} & ${99.78}$ & $98.28$ & $92.83$  \\
  CO-Mining (ICCV'$19$)~\cite{wangco}      & $-$ & $95.87$ & $87.31$ \\
  Arcface (MS$1$M+R$100$)-Official~\cite{deng2018Arcface}\footnote{Results are from the official model:~\url{https://github.com/deepinsight/insightface}, which is trained on MS$1$M and adopts ResNet$100$ as the backbone.} & $\bf{99.82}$ & $98.37$ & $92.08$ \\ 
  Arcface (MS$1$M+R$100$) & $99.80$ & $98.29$ & $92.52$ \\ 
  Arcface (VGG+R$100$) & $99.62$ & $98.30$ & $93.13$ \\ 
  \hline\hline
  \bf Ours (VGG+R$100$)  &  $99.68$ &  $\bf{98.53}$ & $\bf{93.43}$ \\ \hline 
  \end{tabular}
  }
  \end{minipage}
  \tablespace
\end{table}
 \end{savenotes}

\begin{figure}[t!]
  \centering
  \includegraphics[trim={0 0 0 0mm},clip,width=0.92\linewidth]{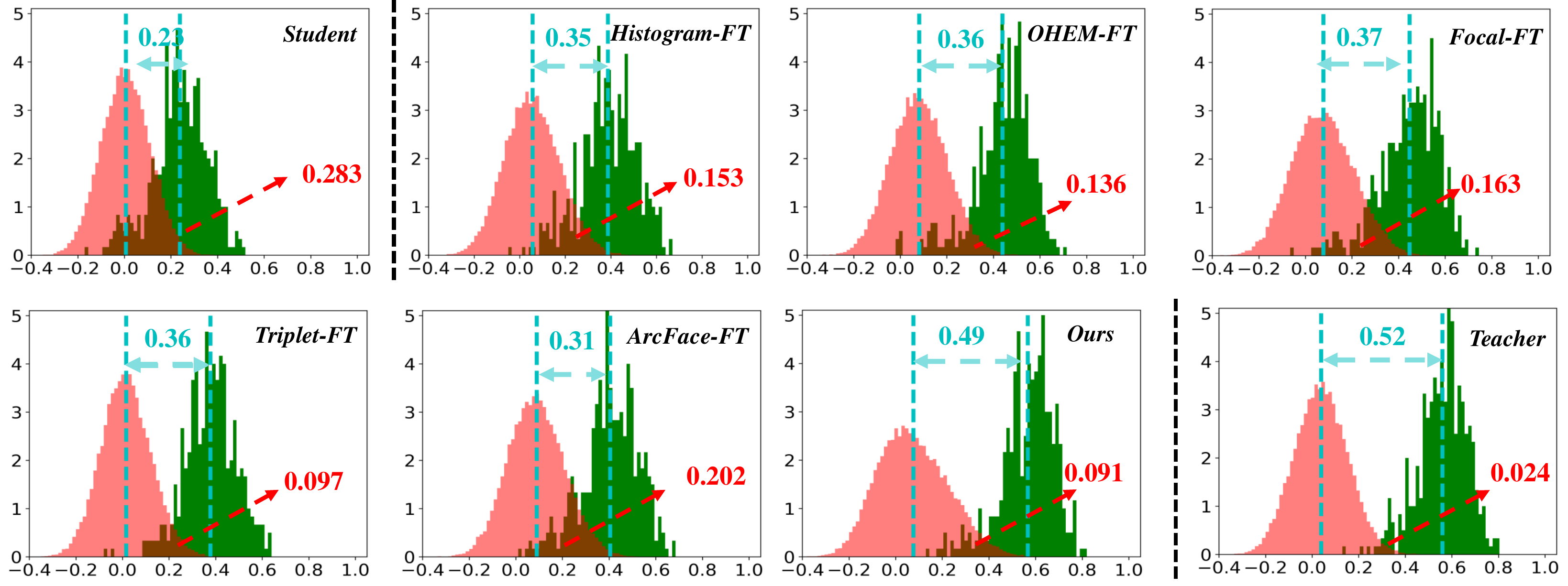}
  \vspace{-3mm}
  \caption{\small \textbf{Illustrations of similarity distributions of different SotA methods}, which are all pre-trained by CASIA with ResNet$50$ and then finetuned on SCface.
  The leftmost and rightmost are the student and teacher distributions estimated from a pre-trained Arcface model on $d_1$ and $d_3$ settings, respectively. 
  The similarity distributions in the middle are obtained by various methods finetuned on SCface. 
  The red number indicates the \textit{histogram intersection} between the estimated similarity distributions from the positive and negative pairs.
  } 
  \label{fig:distribution_compare} \figvspace
\end{figure}

Fig.~\ref{fig:distribution_compare} illustrates the estimated similarity distributions of various SotA methods. 
To quantify the differences among these methods, we introduce two statistics for evaluation, the expectation margin and histogram intersection ($i.e.$, $\sum_{r=1}^{R}\min(h_{r}^{+},h_{r}^{-})$) between the two distributions from positive and negative pairs.
Typically, smaller histogram intersection and larger expectation margin indicate better verification/identification performance, since it means more discriminative embeddings are learned~\cite{Ustinova2016histogram}. 
Our DDL achieves the closest statistics to the teacher distribution, and thus obtains the best performance.

\Paragraph{Pose on CFP-FP and CPLFW.}
We compare our method with SotA pose-invariant methods~\cite{tran2017disentangled,yin2018multi,cao2018pose,deng2018uvgan,peng2017rec} and generic solutions~\cite{wen2016discriminative,liu2017sphereface,cao2018vggface2,deng2018Arcface,wang2018support,wangco}. 
Since VGGFace$2$ includes comprehensive pose variations, we use it to pre-train a ResNet$100$ with Arcface. Next, we construct teacher and student distributions to finetune the model with our loss. 
From Tab.~\ref{tab:comp_pose}, we can see that:
$1$) Our Arcface re-implementations achieve comparable results against the official version,
with similar results on LFW and CFP-FP, as well as better performance on CPLFW.  
Arcface is also much better than other methods, including those pose-invariant face recognition methods.
$2$) Our method achieves the best performance on both pose benchmarks, while also maintaining the performance on LFW (\textit{i.e.}, $99.68\%$ vs. $99.62\%$). 

Note that when using the model pre-trained on MS$1$M, and finetuning it with easy/hard samples from VGGFace$2$, our method can further push the performance to a higher level ($\bf 99.06\%$ on CFP-FP and $\bf 94.20\%$ on CPLFW), which is the \textit{first} method that exceeds $99.0\%$ on CFP-FP and $94.0\%$ on CPLFW using images cropped by MTCNN.
Besides, we also train our DDL on the smaller training set CASIA with a smaller backbone ResNet$50$. 
Again, our DDL outperforms the competitors. 
Please refer to our supplementary material for details.

\Paragraph{Large-Scale Benchmarks: IJB-B and IJB-C.}
On IJB-B/C datasets, we employ VGGFace$2$ with ResNet$50$ for a fair comparison with recent methods.
We first construct the teacher and student distributions according to the pose of each image, and then follow the testing protocol in~\cite{deng2018Arcface} to take the \textit{average of the image features} as the corresponding template representation without bells and whistles. 
Tabs.~\ref{tab:comp_ijb} and~\ref{tab:comp_ijb1vsN} show the $1$:$1$ verification and $1$:N identification comparisons with the recent SotA methods, respectively.
Note that our method is \textit{not} a set-based face recognition method, and the experiments on these two datasets are just to prove that our DDL can obtain more discriminate features than generic methods like Arcface, even on all-variations-included datasets. Please refer to our supplementary material for the detailed analysis.

\begin{table}[t!]
\begin{center}
\small
\caption{\small \textbf{$1$:$1$ verification TAR} on the IJB-B and IJB-C datasets.
All methods are trained on VGGFace$2$ with ResNet$50$.}
\label{tab:comp_ijb}
\resizebox{0.75\linewidth}{!}{
\begin{tabular}{c|ccccccc}
\hlinew{1.2pt}
%\hline
\multirow{2}{*}{Methods  (\%) }        &  \multicolumn{3}{c}{IJB-B} & \multicolumn{3}{c}{IJB-C}\\ 
                         \cmidrule(r){2-4} \cmidrule{5-7}
                         &FAR=$1e$$-$$5$ & FAR=$1e$$-$$4$ & FAR=$1e$$-$$3$ 
                         &FAR=$1e$$-$$5$ & FAR=$1e$$-$$4$ & FAR=$1e$$-$$3$ \\
                         \hline
VGGFace2~\cite{cao2018vggface2}         & $67.1$    & $80.0$ & $88.7$ 
                                                  & $74.7$    & $84.1$ & $90.9$ \\
MN\cite{xie2018multicolumn} & $70.8$ & $83.1$ & $90.9$  
                                                  & $77.1$ & $86.2$ & $92.7$ \\
DCN~\cite{xie2018comparator}          & $-$    & $84.9$ & $93.7$ 
                                                  & $-$    & $88.5$ & $94.7$ \\
Arcface~\cite{deng2018Arcface}        & $80.5$ & $89.9$ & $94.5$ 
                                                  & $86.1$ & $92.1$ & $96.0$ \\
SP~\cite{tung2019sp}                 & $79.4$ & $89.8$ & $94.9$ 
                                                  & $85.9$ &  $92.3$ & $96.2$\\
RKD~\cite{park2019rkd}               & $78.4$ & $89.6$ & $94.7$
                                                  & $85.5$ &  $92.1$& $96.1$ \\
                                                   \hline\hline
\textbf{Ours}  & $\bf{83.4}$ & $\bf{90.7}$ & $\bf{95.2}$ & $\bf{88.4}$ & $\bf{93.1}$ & $\bf{96.3}$ \\\hline
\end{tabular}
}
\end{center}
\tablespace
\end{table}

\begin{table}[t]%\vspace{-1mm}
\begin{center}
\small
%\vspace{-5mm}
\caption{\small \textbf{1:N (mixed media) Identification} on IJB-B/C. All methods are trained on VGGFace$2$ with ResNet$50$. 
VGGFace$2$ is cited from the paper, and Arcface is from its official released model.}
\label{tab:comp_ijb1vsN}
\resizebox{0.9\linewidth}{!}{
% \begin{tabular}{@{}c@{\hspace{1mm}}|@{\hspace{1mm}}c@{\hspace{1mm}}c@{\hspace{1mm}}c@{\hspace{1mm}}c@{\hspace{1mm}}@{\hspace{1mm}}c@{\hspace{1mm}}c@{\hspace{1mm}}c@{\hspace{1mm}}c@{}}
\begin{tabular}{@{}c@{}|@{}c@{}c@{}c@{}c@{}@{}c@{}c@{}c@{}c@{}}
\hlinew{1.2pt}
\multirow{2}{*}{Methods  (\%) }  &  \multicolumn{4}{c}{IJB-B} &  \multicolumn{4}{c}{IJB-C} \\
\cmidrule{2-5} \cmidrule{6-9}
&FPIR=$0.01$ & FPIR=$0.1$ & Rank~$1$ & Rank~$5$ 
&FPIR=$0.01$ & FPIR=$0.1$ & Rank~$1$ & Rank~$5$ \\\hline
VGGFace2~\cite{cao2018vggface2}         & $70.6$   & $83.9$ & $90.1$  & $94.5$ 
                 & $74.6$   & $84.2$ & $91.2$  & $94.9$ \\
Arcface ~\cite{deng2018Arcface}          & $73.1$   & $88.2$ & $93.6$  & $96.5$ 
                 & $79.6$   & $89.5$ & $94.8$  & $96.9$ \\
SP~\cite{tung2019sp}      &   $72.4$ &  $88.0$  &    $93.8$   & $96.6$  
                          &   $79.9$ &  $89.5$  &    $94.7$   & $97.0$  \\
RKD~\cite{park2019rkd}      &   $70.6$ & $87.6$   &  $93.4$ & $96.5$
                            &   $79.3$ & $89.1$   &  $94.6$ & $96.9$    \\ \hline\hline
Ours             & $\mathbf{76.3}$   & $\mathbf{89.5}$ & $\mathbf{93.9}$  & $\mathbf{96.6}$
                 & $\mathbf{85.4}$   & $\mathbf{91.1}$ & $\mathbf{95.4}$  & $\mathbf{97.2}$ 
\\\hline 
\end{tabular} 
}
%\vspace{-5mm}
\end{center}
\end{table}

\Paragraph{Comparisons with SotA KD Methods.} 
We further conduct fair comparisons between our DDL and the recent SotA KD/self-distillation methods, \textit{i.e.}, SP~\cite{tung2019sp} and RKD~\cite{park2019rkd}.
Note that since both SP and RKD have not reported SOTA results on face recognition tasks, we re-implement the two methods under the same experimental setting on VGGFace$2$, using their officially released code.
Specifically, we first train a ResNet$50$ with Arcface on VGGFace$2$ as the teacher model, and then train a student ResNet$50$ via combining the knowledge distillation method (\textit{e.g.}, SP or RKD) and Arcface loss under the guidance of the teacher model. 
As in Tabs.~\ref{tab:comp_ijb} and~\ref{tab:comp_ijb1vsN}, 
our DDL outperforms the SotA KD/self-distillation methods, which achieve similar results to vanilla Arcface. 

\section{Conclusion}
In this paper, we propose a novel framework Distribution Distillation Loss (DDL) to improve various variation-\textit{specific} tasks, which comes from the observations that state-of-the-art methods (\textit{e.g.}, Arcface) witness significant performance gaps between easy and hard samples.
The key idea of our method is to construct a teacher and a student distribution from easy and hard samples, respectively.
Then, the proposed loss drives the student distribution to approximate the teacher distribution to reduce the overlap between the positive and negative pairs.
Extensive experiments demonstrate the effectiveness of our DDL on a wide range of recognition tasks compared to the state-of-the-art face recognition methods.
In subsequent work, we can try to extend our method to multiple teacher-student distribution pairs for the corresponding task respectively. 
%Besides, motivated by the idea of online hard sample mining, DDL can be easily extended to an online version that automatically defines the easy/hard samples regardless of the type of variations, which can also be a good future direction. 

\clearpage
\bibliographystyle{splncs04}
\bibliography{egbib}
\end{document}

% \linenumbers
\pagestyle{headings}
\mainmatter
\def\ECCVSubNumber{7177}  % Insert your submission number here

\title{Supplementary Material for "Improving Face Recognition from Hard Samples via Distribution Distillation Loss"} % Replace with your title

\begin{comment}
\titlerunning{ECCV-20 submission ID \ECCVSubNumber} 
\authorrunning{ECCV-20 submission ID \ECCVSubNumber} 
\author{Anonymous ECCV submission}
\institute{Paper ID \ECCVSubNumber}
\end{comment}
%******************

%\begin{comment}
\titlerunning{Distribution Distillation Loss}
\author{First Author\inst{1}\orcidID{0000-1111-2222-3333} \and
Second Author\inst{2,3}\orcidID{1111-2222-3333-4444} \and
Third Author\inst{3}\orcidID{2222--3333-4444-5555}}
\authorrunning{Y. Huang, P. Shen, and et al.}
\author{Yuge Huang${}^\ast$\inst{1} \and Pengcheng Shen${}^\ast$\inst{1} \and Ying Tai${}^\sharp$\inst{1} \and Shaoxin Li${}^\sharp$\inst{1} \and Xiaoming Liu\inst{2} \and Jilin Li\inst{1} \and Feiyue Huang\inst{1} \and Rongrong Ji\inst{3}}
\authorrunning{Y. Huang, P. cheng, and et al.}
\institute{Youtu Lab, Tencent \and
Michigan State University \and
Xiamen University\\
\email{\{yugehuang, quantshen, yingtai, darwinli, jerolinli, garyhuang\}@tencent.com, liuxm@cse.msu.edu, rrji@xmu.seu.cn}}
%\end{comment}
%******************
\maketitle

%\maketitle
%\thispagestyle{empty}

This supplementary material provides additional details on the following:
%\vspace{-2mm}
\begin{tight_itemize}
\item Gradient derivation of Eq.~$3$ (Paper).
\item T-SNE illustrations on feature distribution in different iterations.
\item Effects on Numbers of \textit{b} and Bins.
\item Experiments trained by CASIA and tested on CFP-FP with ResNet$50$ as the backbone.
\item Comparisons on IJB-B and IJB-C between ArcFace and our DDL.
\item A simple attempt of online DDL compared to the offline version in our submission.
\item Time Complexity of the method.
\end{tight_itemize}

%{\small
%\bibliographystyle{ieee}
%\bibliography{egbib}
%}

\begin{figure}[t]
    \centering
    \includegraphics[trim={0 0 0 0mm},clip,width=1\linewidth]{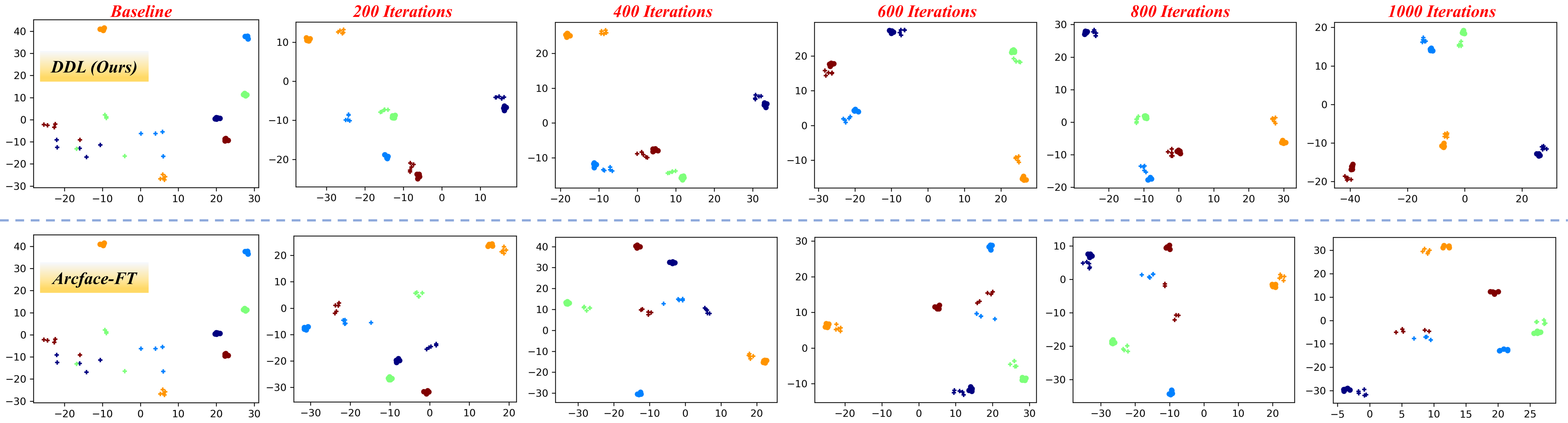}
    %\vspace{-7mm}
    \captionof{figure}{\small \textbf{T-SNE illustration on feature distribution} in different training iterations on SCface.
    \textbf{Top}: Ours.
    \textbf{Bottom}: Arcface-FT.
    The same color indicates samples of the same subject.
    `\textbf{$\bullet$}' indicates samples captured with the Distance3 (\textit{i.e.}, $d_3$) setting, and `$\textbf{+}$' is samples with Distance1 (\textit{i.e.}, $d_1$).
    $d_1$ and $d_3$ indicate images were captured at distances of $4.2$ and $1.0$m, respectively.
    Note that because of the characteristic of T-SNE algorithm, the axes of the same subject may keep changing.
    Please zoom in to see more details.}
    \label{fig:t-sne} 
\end{figure}

\section{Gradient Derivation of Eq.~$3$ (Paper)}
First, we present the detailed derivation of Eq.~$3$ in our submission. 
As described in Sec.~$3.1$ (Paper), we have $h_{r}^{+}$ of the histogram $H^{+}$ at each bin as: 
\begin{equation}
    h_{r}^{+}=\frac{1}{|S^+|}\sum_{(i,j):m_{ij=+1}}\delta_{i,j,r},
\end{equation}
where $\delta_{i,j,r}$ is the weight and is defined as:
\begin{equation}
    \delta_{i,j,r} = \text{exp}(-\gamma{(s_{ij} - t_{r})^{2}}).
\end{equation}
For any $s_{ij}$,
$\frac{\partial h_{r}^{+}}{\partial s_{ij}}$ can be obtained as follows:
\begin{equation}
    \frac{\partial h_{r}^{+}}{\partial s_{ij}} = \frac{-2 \cdot\gamma \cdot \delta_{i,j,r}(s_{ij} - t_{r})}{|S^+|}.
\end{equation}

\begin{figure}[t!]
  \centering
  \includegraphics[trim={0 4 0 0mm},clip,width=0.98\linewidth]{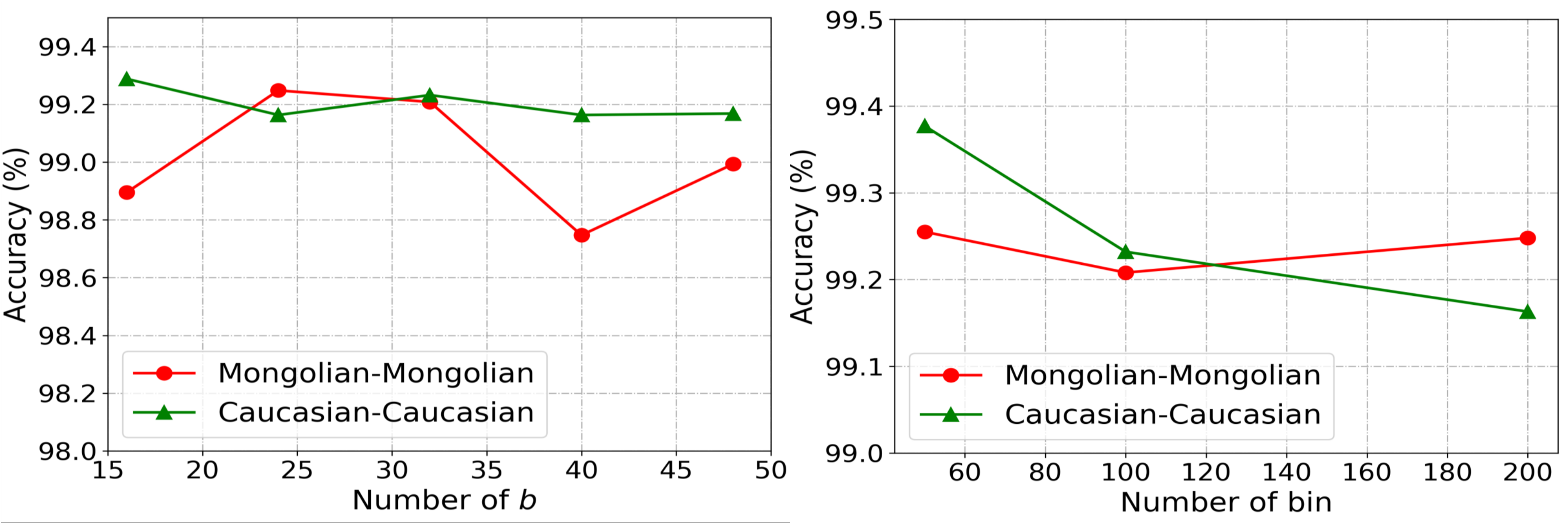}
  %\vspace{-3mm}
  \caption{\small \textbf{Effects on Numbers of \textit{b} and bins} on COX datasets.
  Our loss is not sensitive to the batch size or the number of bins.
  } 
  \label{fig:batch_bin} \figvspace
\end{figure}

\section{T-SNE Illustrations on Feature Distribution in Different Iterations}
As shown in Fig.~\ref{fig:t-sne}, we compare Arcface-FT and our method, and illustrate the T-SNE visualizations of feature distribution in different training iterations. 
We randomly select $5$ subjects in the test set, and the baseline indicates the original Arcface model.
It is shown that:
$1$) The original Arcface performs well on easy samples (\textit{i.e.}, $d_3$ setting), but poorly on hard samples (\textit{i.e.}, $d_1$ setting).
$2$) After finetuning, Arcface-FT significantly improves the performance on some hard samples, but still fails on some subjects.
$3$) Compared to Arcface-FT, our method better addresses the hard samples, which makes the easy/hard samples from the same subject more \textit{compact} in the feature space.

\Section{Effects on Numbers of \textit{b} and Bins}
Since the similarity distributions of positive and negative pairs are estimated within a mini-batch, we conduct tests on COX to explore the effect of the number of \textit{b} in Fig.~\ref{fig:batch_bin}, and observe that the results remain stable when \textit{b} varies from $16$ to $48$ along with the number of bins equals to $100$. 
In addition, we also investigate how the number of bins affects performance by fixing the number of \textit{b} to be $32$. 
As shown in Fig.~\ref{fig:batch_bin}, the results also remain stable. 
To sum up, our loss is not sensitive to the numbers of \textit{b} and bins. 

\begin{figure}[t!]
  \centering
  \includegraphics[trim={0 0 0 0mm},clip,width=0.6\linewidth]{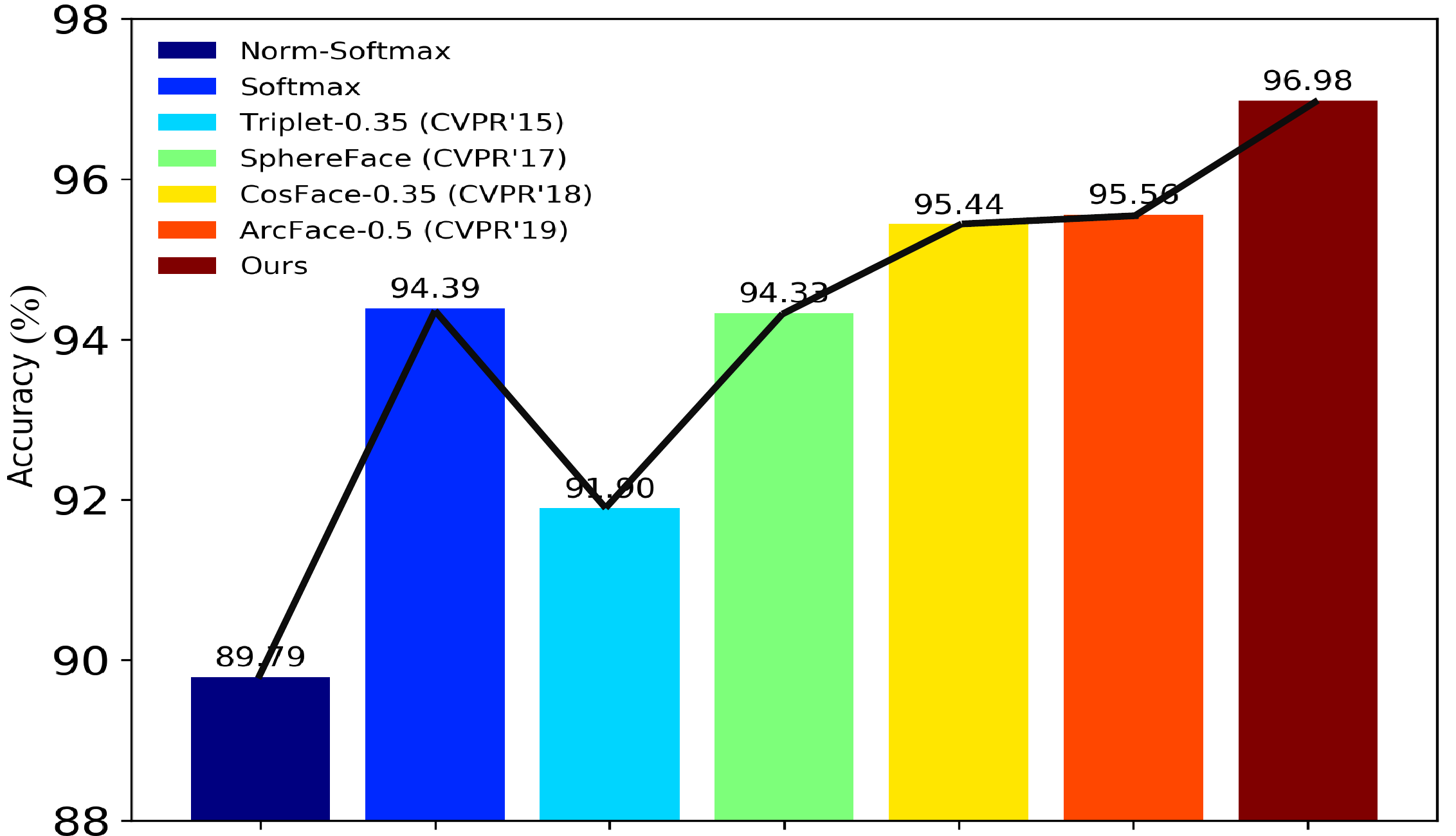}
  %\vspace{-1mm}
  \caption{\small \textbf{Verification results of different loss functions} on CFP-FP. 
  All methods are trained by dataset CASIA with ResNet$50$ as the backbone.
  }
  \label{fig:comp_loss} \figvspace
\end{figure}

\begin{figure}[t!]
\centering
\subfigure[ROC for IJB-B]{
\begin{minipage}[t]{0.49\linewidth}
\centering
\label{fig:ijbb_roc}
\includegraphics[width=1\linewidth]{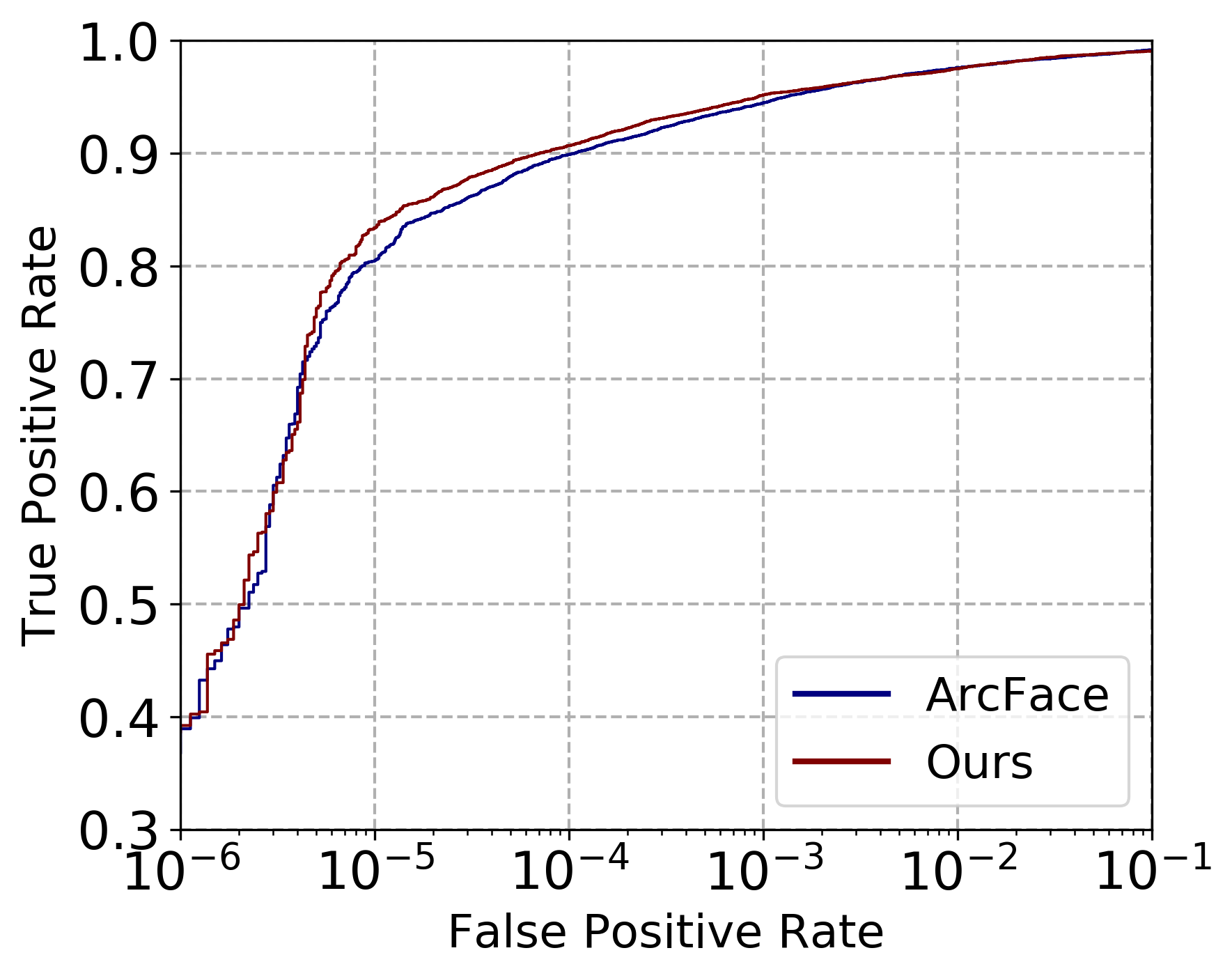}
%\caption{fig1}
\end{minipage}%
}%
\subfigure[ROC for IJB-C]{
\begin{minipage}[t]{0.49\linewidth}
\centering
\label{fig:ijbc_roc}
\includegraphics[width=1\linewidth]{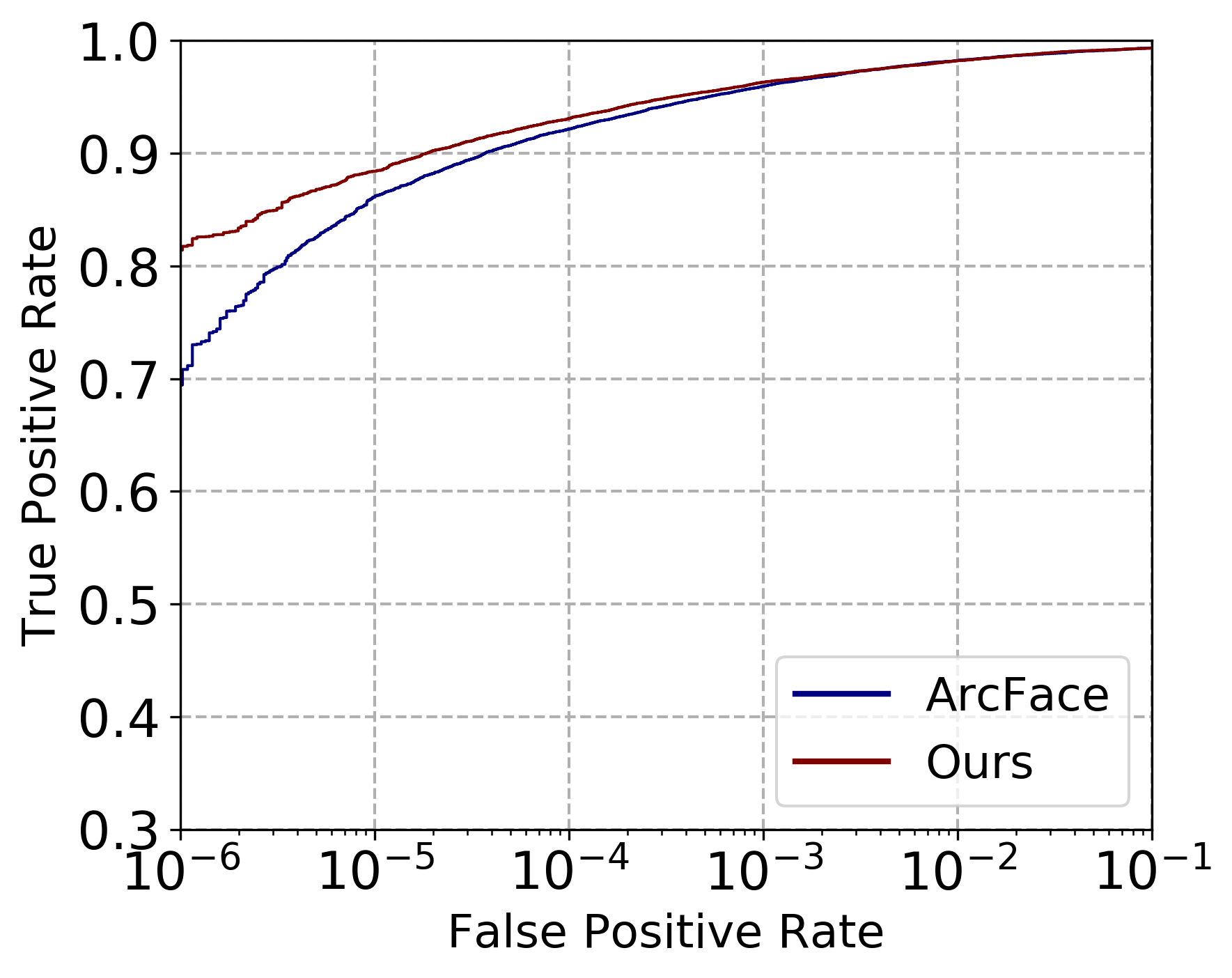}
%\caption{fig2}
\end{minipage}%
}%
\centering
%\vspace{-4mm}
\caption{\small ROC curves of $1$:$1$ verification protocol on the IJB-B and IJB-C datasets. Both ArcFace and ours are trained by VGGFace$2$ dataset with backbone ResNet$50$.
}
 \label{fig:ijb_roc}
\end{figure}

\section{Pose on CFP-FP}
 We compare our loss with other representative losses in face recognition, including Softmax, Norm-Softmax, Triplet loss, SphereFace, CosFace and ArcFace on CFP-FP dataset.
All of the methods are trained by the same dataset CASIA with the same backbone ResNet$50$.
For our method, we estimate the pose of each image in CASIA and construct teacher (yaw$<10^{\circ}$) and student distributions (yaw$>45^{\circ}$), respectively.
Fig.~\ref{fig:comp_loss} indicates that our loss not only performs better than the triplet-based method (\textit{i.e.}, Triplet, $91.9\%$), but also better than the softmax-based methods (\textit{i.e.}, Arcface, $95.6\%$).

\section{Comparisons on IJB-B and IJB-C}
\begin{figure}[t!]
\centering
\subfigure[]{
\begin{minipage}[t]{0.5\linewidth}
\centering
\label{fig:ijb_example1}
\includegraphics[width=0.98\linewidth]{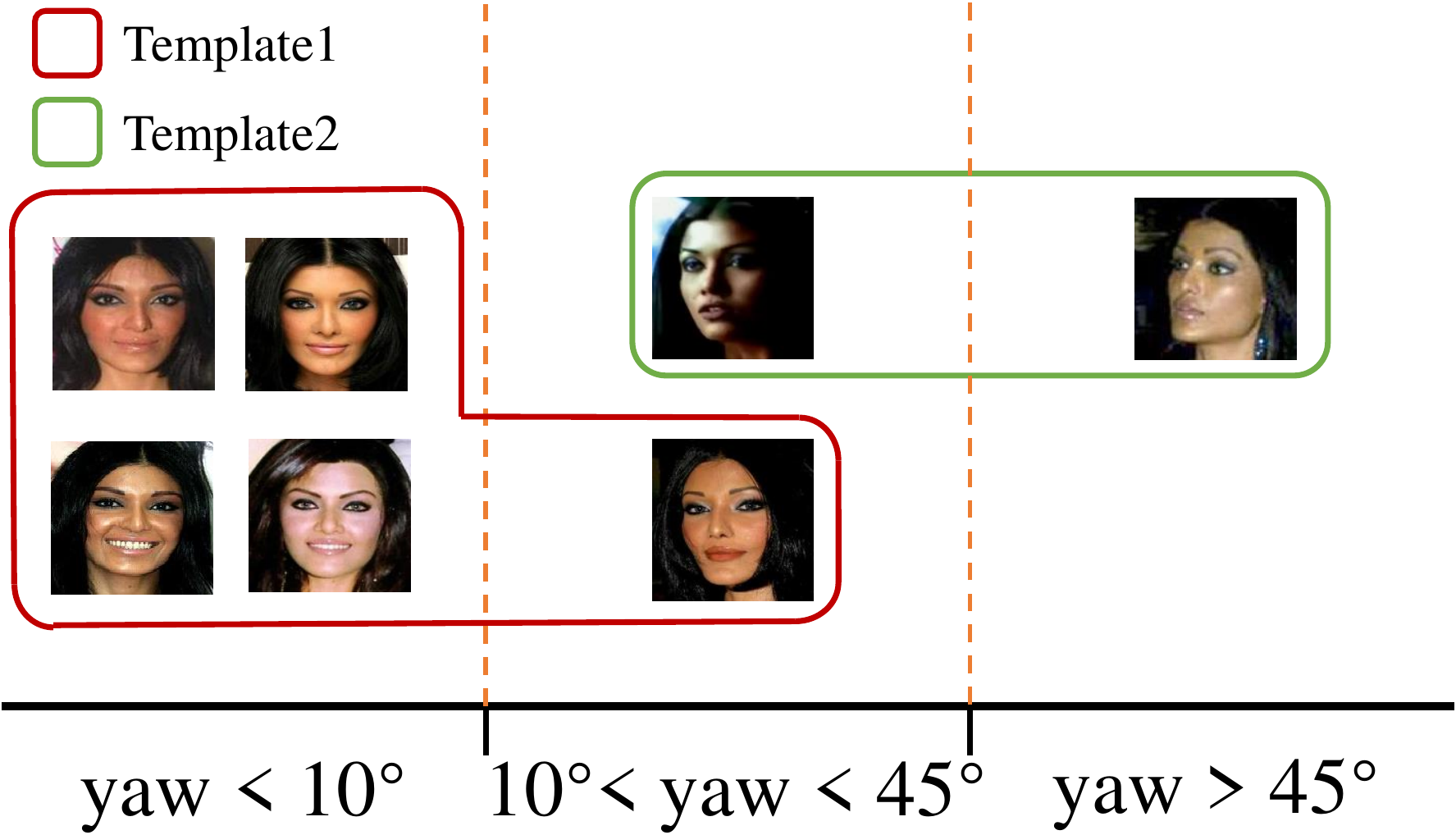}
%\caption{fig1}
\end{minipage}%
}%
\subfigure[]{
\begin{minipage}[t]{0.5\linewidth}
\centering
\label{fig:ijb_example2}
\includegraphics[width=0.98\linewidth]{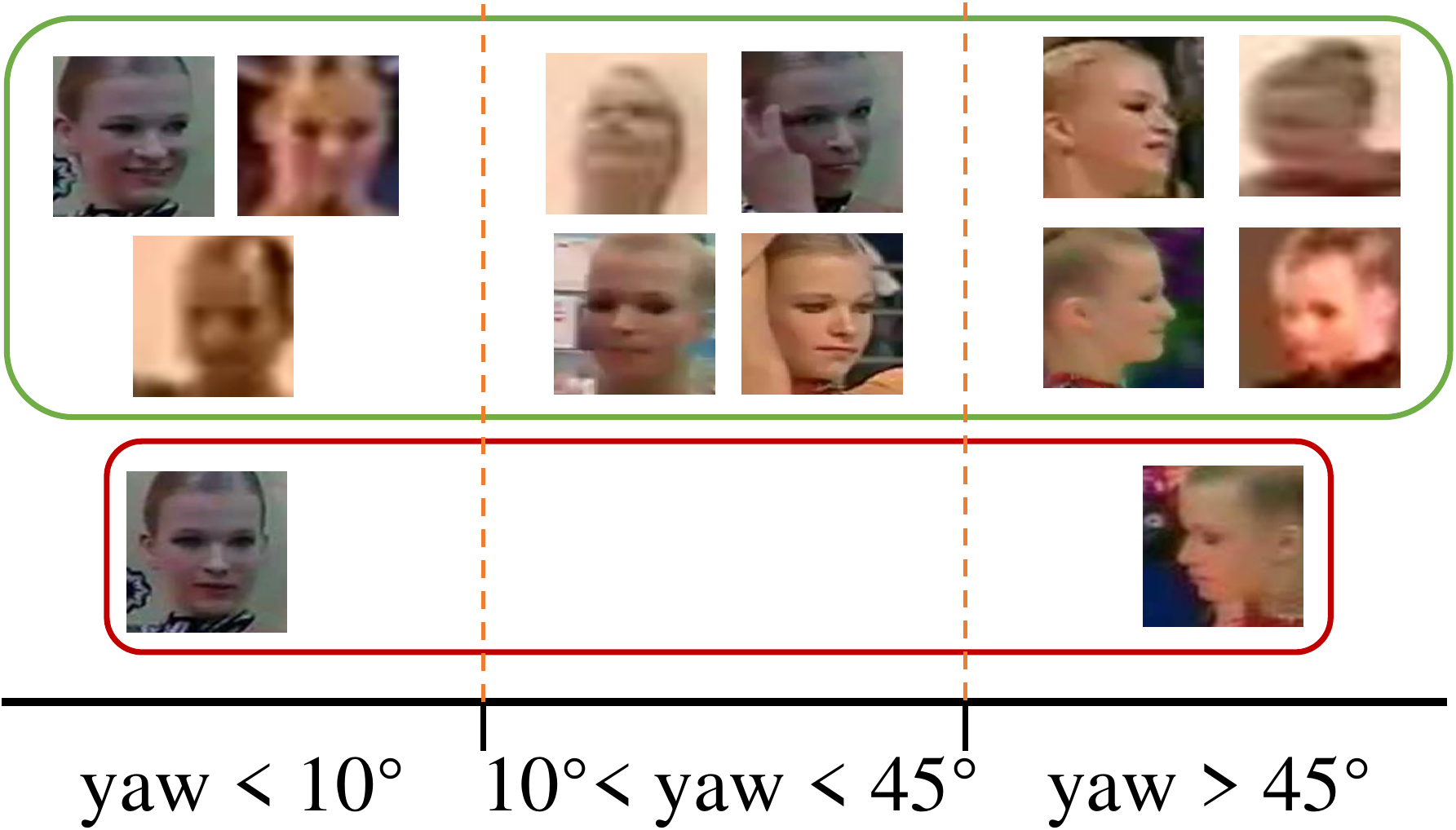}
%\caption{fig2}
\end{minipage}%
}%
\centering
%\vspace{-4mm}
\caption{\small \textbf{Examples of positive template pairs classified correctly by ours while ArcFace fails} at FAR=$1$$e$$-5$ from IJB-B.
  }
 \label{fig:ijb_example}
\end{figure}

Fig.~\ref{fig:ijb_roc} shows the ROC curves of our DDL and ArcFace on IJB-B and IJB-C datasets.
As we can see, our method achieves better performance and forms an upper envelope compared to ArcFace~\footnote{Results are from the official ResNet$100$ pre-trained with MS$1$M:~\url{https://github.com/deepinsight/insightface}.}.
Fig.~\ref{fig:ijb_example} illustrates two examples of positive template pairs that are classified correctly by ours but Arcface fails. 
Benefiting from the distribution constraint on pose variation, our DDL learns more \textit{discriminant pose-invariant features}, and thus leads to better performance on templates that differ on poses (Fig.~\ref{fig:ijb_example1}).
Interestingly, when comprehensive variations exist, \textit{e.g.}, resolution (Fig.~\ref{fig:ijb_example2}), Arcface again fails while our method shows robustness to some extent.

\begin{table}[t!]
\begin{center}
\small
\caption{\small \textbf{Comparisons among Arcface and two versions of DDL}.
We choose VGGFace$2$ as the large scale dataset and test on IJB-B dataset, while CASIA is adopted as a small scale dataset and CFP-FP is tested.}
\label{tab:comp_ijb}
\resizebox{1\linewidth}{!}{
\begin{tabular}{c|ccc|cccc|c}
\hlinew{1.2pt}
%\hline
\multirow{2}{*}{Methods  (\%) }        &  \multicolumn{3}{c}{IJB-B(1:1)} & \multicolumn{4}{c}{IJB-B(1:N)} & \multicolumn{1}{c}{CFP-FP}\\ 
                         \cmidrule(r){2-4} \cmidrule{5-8} \cmidrule{9-9}
                         &FAR=$1e$$-$$5$ & FAR=$1e$$-$$4$ & FAR=$1e$$-$$3$ 
                         &FPIR=$0.01$ & FPIR=$0.1$ & Rank1 & Rank5 & Acc\\
                         \hline
Arcface       & $80.5$ & $89.9$ & $94.5$  & $73.1$ & $88.2$ & $93.6$ & $96.5$ &$95.58$\\
Online DDL    & $81.5$ & $90.7$ & $95.1$  & $75.1$ & $89.0$ & $93.6$ & $96.6$ &$96.82$\\
Offline DDL   & $83.4$ & $90.7$ & $95.2$  & $76.3$ &  $89.5$& $93.9$ & $96.6$ &$96.98$\\\hline
\end{tabular}
}
\end{center}
\tablespace
\end{table}

\Section{Simple Attempt of Online DDL}
In our submission, the hard samples are defined by the variation of the image, such as pose, resolution, and race (\textit{i.e.}, offline DDL). Here, we simply explore an online version of DDL, which mines the hard samples during training. 
Specifically, in each iteration, we randomly select $64$ classes, in which $8$ samples are randomly picked from each class and thus construct a mini-batch with $512$ samples.
We then sort the samples from the same class with losses during training. 

In our online DDL, \textit{the first $4$ samples with larger losses are defined as hard samples, and the others are easy samples}. 
We conduct the experiments on both small and large-scale training datasets, \textit{i.e.,} CASIA and VGGFace2, respectively. 
As shown in Tab.~\ref{tab:comp_ijb}, our simple attempt of online DDL is slightly inferior to the offline DDL in our submission, but is still superior to ArcFace, which demonstrates the effectiveness of the idea of DDL, even when easy and hard samples are defined in different ways.
The nature of our method lies in the distillation between two similarity distributions constructed from easy and hard samples, it's also interesting future direction to find out a better solution to define easy and hard samples.
%shows the feasibility of such strategy. 

\Section{Time Complexity}
Compared with conventional finetuning, the only additional operation in our training is to construct the positive and negative pairs in each mini-batch. 
The positive pairs are constructed offline, and thus has little influence on the training speed.
For negative pairs, since we adopt online hard sample mining to grab the hardest $b$ pairs from the total $b\times (b-1)/2$ negative pairs for each distribution, our training step is slightly slower than conventional finetuning, which is still fast. 
Specifically, with the same environment, conventional finetuning on COX dataset costs $39$ minutes for training, while ours costs $43$ minutes.
Hence, our DDL only has small effects on the training process and has \textbf{NO} influence on inference.

\clearpage
%\bibliographystyle{splncs04}
%\bibliography{egbib}